\documentclass{article}

\usepackage{iclr2027_conference,times}

\usepackage{amsmath,amssymb,amsthm,mathtools}
\usepackage{bm}
\usepackage{booktabs}
\usepackage{multirow}
\usepackage{graphicx}
\usepackage{algorithm}
\usepackage[noend]{algpseudocode}
\usepackage{microtype}
\usepackage{enumitem}
\usepackage{xcolor}
\usepackage{tikz}
\usepackage{wrapfig}
\usetikzlibrary{arrows.meta,positioning,fit,calc}
\usepackage[hidelinks]{hyperref}
\usepackage{url}
\usepackage[capitalize,noabbrev]{cleveref}
\usepackage{placeins}
\makeatletter
\providecommand{\theHALG@line}{}
\renewcommand{\theHALG@line}{\thealgorithm.\arabic{ALG@line}}
\makeatother

\usepackage{amsmath,amsfonts,bm}
\usepackage{lipsum}                     % Dummytext
\usepackage{xargs}
\usepackage{xfrac}
\usepackage{amssymb}
\usepackage{amsmath}
\usepackage{bm} 
\usepackage{mathtools} 
\usepackage{amsthm}
\usepackage{mleftright}
\usepackage{refcount}

\def\1{\bm{1}}

\DeclareMathAlphabet{\mathsfit}{\encodingdefault}{\sfdefault}{m}{sl}
\SetMathAlphabet{\mathsfit}{bold}{\encodingdefault}{\sfdefault}{bx}{n}

\newcommand{\E}{\mathbb{E}}

\newcommand{\R}{\mathbb{R}}

\DeclareMathOperator*{\argmax}{arg\,max}
\DeclareMathOperator*{\argmin}{arg\,min}

\newtheorem{theorem}{Theorem}

\newtheorem{proposition}[theorem]{Proposition}

\theoremstyle{definition}

\theoremstyle{remark}

\crefname{algorithm}{algorithm}{algorithms}
\Crefname{algorithm}{Algorithm}{Algorithms}
\crefname{theorem}{theorem}{theorems}
\Crefname{theorem}{Theorem}{Theorems}
\crefname{lemma}{lemma}{lemmas}
\Crefname{lemma}{Lemma}{Lemmas}
\crefname{proposition}{proposition}{propositions}
\Crefname{proposition}{Proposition}{Propositions}
\crefname{corollary}{corollary}{corollaries}
\Crefname{corollary}{Corollary}{Corollaries}
\crefname{assumption}{assumption}{assumptions}
\Crefname{assumption}{Assumption}{Assumptions}

\providecommand{\R}{\mathbb{R}}
\providecommand{\E}{\mathbb{E}}

\newtheorem*{repeatthminner}{\repeatthmname}
\newcommand{\repeatthmname}{}

\newtheorem*{repeatlemmainner}{\repeatlemmaname}
\newcommand{\repeatlemmaname}{}

\newtheorem*{repeatpropinner}{\repeatpropname}
\newcommand{\repeatpropname}{}

\newtheorem*{repeatcorollaryinner}{\repeatcorollaryname}
\newcommand{\repeatcorollaryname}{}

\def\balpha{\boldsymbol\alpha}

\def\bq{\mathbf{q}}
\def\bv{\mathbf{v}}
\def\bx{\mathbf{x}}
\def\bu{\mathbf{u}}

\def\bz{\mathbf{z}}

\def\E{\mathbb{E}}

\newcommand{\mc}[1]{\mathcal{#1}}

\theoremstyle{plain}
\newtheorem{myremark}[theorem]{Remark}

\definecolor{myOrange}{RGB}{235, 105, 92}
\definecolor{myMaroon}{RGB}{128, 0, 0}
\definecolor{maroon}{rgb}{0.5,0.0,0.0}
\definecolor{heavymaroon}{rgb}{0.3,0.0,0.0}

\definecolor{mygreen}{RGB}{160, 200, 140}

\hypersetup{ 
colorlinks=true, 
linkcolor=myOrange, 
citecolor=myMaroon, filecolor=cyan, 
urlcolor=mygreen
}

\definecolor{Red}{RGB}{244,124,124}
\definecolor{Green}{RGB}{162,222,147}
\definecolor{Blue}{RGB}{112,161,211}
\providecommand{\rkone}[1]{{\setlength{\fboxsep}{1.5pt}\colorbox{Red!50}{#1}}}
\providecommand{\rktwo}[1]{{\setlength{\fboxsep}{1.5pt}\colorbox{Blue!50}{#1}}}

\usepackage{xcolor}
\newcommand{\rev}[1]{\textcolor{black}{#1}}

\title{\rev{Fast Neural Combinatorial Optimization via Parallel Oracles}}

\newcommand{\authorname}[1]{%
    \makebox[\dimexpr0.5\textwidth-2\tabcolsep\relax][l]{\textbf{#1}}%
}

\author{\authorname{Ashkan Soleymani\thanks{Equal contribution.}
 ~\thanks{Part of this work was done during an internship at RBC Borealis.}} \\
Massachusetts Institute of Technology \\
\texttt{ashkanso@mit.edu}
\And
\authorname{Yifei Xu\footnotemark[1]} \\
Texas A\&M University \\
\texttt{yifei\_xu@tamu.edu}
\AND
\authorname{Nikolaos Karalias} \\
Massachusetts Institute of Technology \\
\texttt{stalence@mit.edu}
\And
\authorname{Akbar Rafiey} \\
RBC Borealis, New York University \\
\texttt{akbar.rafiey@rbc.com}
}

\iclrfinalcopy

\begin{document}
\maketitle

\begin{abstract}
Self-supervised learning for combinatorial optimization offers a promising way to train neural solvers without requiring optimal solutions as supervision. This requires informative gradients while respecting discrete feasibility. Recent geometric approaches use Carath\'eodory decompositions to decompose a neural network output into a distribution over feasible solutions, but the resulting decomposition procedures can be sequential and computationally expensive. We propose a parallel alternative based on linear optimization oracles. A neural network generates multiple query directions, whose samples are passed independently through the oracle to produce feasible solutions in parallel. We then obtain a probability distribution over these feasible solutions through a regularized simplex-constrained quadratic program and use its KKT conditions to derive the gradients of the designed loss, without differentiating through the discrete oracle. The resulting method relies mainly on parallel oracle calls and GPU-friendly linear algebra. Experiments across constrained combinatorial optimization problems show that it produces high-quality feasible solutions while substantially reducing the computational cost of decomposition-based self-supervised learning.
\end{abstract}

\vspace{-0.2cm}
\section{Introduction}
\vspace{-0.1cm}
Neural combinatorial optimization (NCO) aims to learn algorithms for
solving combinatorial optimization problems from distributions of problem
instances. A central challenge is to design neural solvers that are both
broadly applicable across different combinatorial structures and
computationally competitive with specialized solvers and heuristics. While
neural networks naturally benefit from massively parallel GPU computation,
the surrounding optimization pipeline---including constraint handling,
solution recovery, and other combinatorial subroutines---may not be equally
well suited to the same hardware. As a result, these components can dominate
runtime and limit the scalability of the overall solver. This motivates designing NCO pipelines whose training and solution-generation
steps can both exploit GPU parallelism.

Scalability also concerns the cost of supervision. Neural methods typically learn from large collections of problem instances, and procuring high-quality solution labels for these instances can require expensive calls to combinatorial solvers. Self-supervised NCO offers an appealing direction because it uses the combinatorial objective itself as the training signal, avoiding the need to generate labels of optimal or near-optimal solutions. \rev{Combined with GPU-parallel algorithms, this paradigm offers a route to computationally scalable neural solvers without requiring costly solution labels.}

% Combined with algorithms that exploit GPU parallelism, this paradigm offers a route to neural solvers that scale both computationally and with respect to the cost of supervision.

A central challenge in self-supervised constrained NCO is how to incorporate
discrete feasibility constraints into a differentiable learning pipeline. A
common approach is to optimize a continuous surrogate while penalizing
constraint violations in the loss. Other approaches rely on continuous
relaxations, differentiable projections, or problem-specific layers that map
the network output toward the feasible region
\citep{karalias2020erdos,wang2022unsupervised,donti2021dc3,
wang2023linsatnet,zeng2024glinsat}.
While effective in many settings, these methods may require problem-specific
loss design, additional hyperparameter tuning, or a separate rounding or
solution-recovery procedure to obtain a feasible discrete solution. \rev{
Reinforcement learning (RL) is another widely used label-free approach to NCO \citep{bello2017neural,kool2019attention,
kwon2020pomo}. Many RL-based methods rely on policy-gradient estimators, which can suffer from high variance, and handle feasibility through problem-specific action masking or construction rules \citep{nazari2018reinforcement,kool2019attention}.
} Thus,
the choices made for constraint handling, loss computation, and solution
recovery can substantially affect the efficiency and utility of the overall solver.

A recent line of work connects these components through \emph{extensions} of
the discrete objective constructed from the geometry of the feasible set
\citep{Karalias2022NeuralSF,karalias2025geometric,
nerem2025differentiable,Rafiey2026-FWNCO}.
These methods represent, or approximate, a continuous point using weighted
combinations of feasible discrete solutions. The resulting weights define a
distribution over feasible solutions and induce a continuous surrogate through
the expected value of the discrete objective. Under suitable regularity
conditions, this surrogate can be differentiated and used directly for
self-supervised training. Crucially, the same decomposition also produces
feasible candidates for discrete inference. Extensions are therefore
algorithmically appealing because a single geometric construction addresses
both loss design and solution recovery. However, existing geometric and
permutation-based constructions generate these feasible solutions and distributions through
sequential decomposition procedures, creating a computational bottleneck
shared by training and inference.

For constrained NCO, \cite{karalias2025geometric} develop a general framework
based on geometric decompositions of the feasible-set polytope. Their generic
construction builds on the classical Gr\"otschel--Lov\'asz--Schrijver (GLS)
framework, while more efficient variants exploit problem-specific structure.
For permutation problems, \cite{nerem2025differentiable} construct a related
extension through successive decompositions over the Birkhoff polytope.
More recently, \cite{Rafiey2026-FWNCO} propose a simpler oracle-based
Frank--Wolfe decomposition that requires only one linear optimization oracle (a.k.a LMO) call per iteration and applies across a broad class of polytopes. Despite these
improvements, all of these constructions remain \emph{sequential}: the feasible
solution generated at one step requires calling a combinatorial LMO and influences the point or residual used to
generate the next.

This sequential dependence can become a computational bottleneck in neural
optimization pipelines. Even when an individual oracle call is efficient,
the calls cannot be executed simultaneously, and repeated CPU--GPU
communication can introduce additional overhead. Since the same decomposition
is used during both training and inference, removing this dependency could
improve both stages. This motivates our central question: 
\begin{quote}
    \emph{Can we design parallel decomposition algorithms for self-supervised constrained NCO that make both loss computation and rounding efficient on GPUs?}
\end{quote}
% \emph{Can we design parallel decomposition algorithms for self-supervised constrained NCO that make both loss computation and rounding efficient on GPUs?}
More specifically, can we replace long sequences of dependent oracle calls with a small number of batched computations that generate multiple feasible candidates in parallel, while retaining feasibility, differentiability, and the principled connection between continuous training and discrete inference provided by geometric extensions?

\begin{figure}
    \centering
\includegraphics[width=1\linewidth]{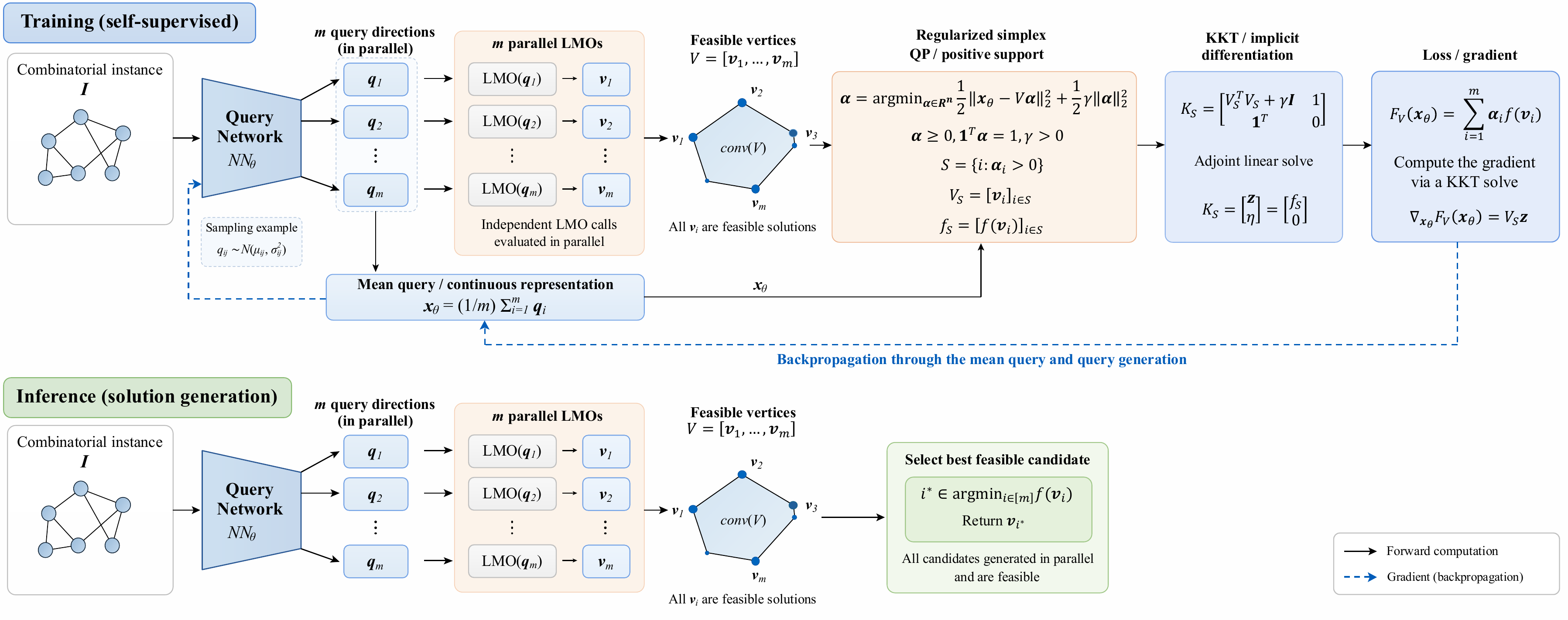}
    \caption{Overview of our approach. The query network produces $m$ query directions, and independent LMOs generate feasible vertices $V$ in parallel. During training, the mean query $\bx_\theta$ is mapped to a convex combination of $V$ via the regularized simplex QP and differentiated through its KKT system. At inference, we return the best feasible vertex among the $m$ candidates.}
    \label{fig:overall-pipline}
    \vspace{-0.5cm}
\end{figure}

\vspace{-0.1cm}
\noindent\textbf{Our contributions.} We address this bottleneck through parallel oracle-based candidate
generation, illustrated in Figure~\ref{fig:overall-pipline}.
Given a problem instance, a neural network generates multiple query
directions, and independent oracle calls map these directions to
feasible combinatorial solutions. During training, a geometric
extension over the generated candidates turns their objective
values into a gradient for updating the network. At inference,
the same candidate-generation procedure is used, and we simply
return the candidate with the best objective value.
Thus, feasibility is enforced by the oracle, while training and
inference share a parallel solution-generation mechanism. The framework accommodates both exact and approximate linear
optimization oracles. Its central technical contribution is a
regularized geometric loss whose gradient is derived from
the associated KKT conditions, enabling training without
differentiating through the discrete oracle. \rev{Our framework works across diverse combinatorial optimization problems with different constraints and polytope structures.} We evaluate it on Maximum Coverage, Quadratic Assignment,
MaxCut, and Densest $k$-Subgraph, demonstrating competitive solution
quality and low inference cost across different combinatorial
structures, including graphs with millions of vertices.

% \paragraph{Our contributions.}
% We make the following contributions:
% \begin{itemize}[leftmargin=*]
%     \item \textbf{Parallel oracle-based decomposition.}
%     We introduce a self-supervised NCO framework that replaces sequential
%     geometric decompositions with independent linear optimization oracle
%     calls. A neural network generates multiple query distributions whose
%     samples are processed in parallel to produce feasible combinatorial
%     solutions.

%     \item \textbf{Differentiable geometric extension.}
%     We construct an extension over the generated feasible solutions and
%     derive its gradient from the associated KKT system. This yields a simple
%     backward computation without differentiating through the discrete oracle.

%     \item \textbf{Unified training and inference.}
%     The same feasible solutions used to define the training objective are
%     used directly as candidates at inference time, avoiding a separate
%     solution-recovery procedure. The framework only requires access to an
%     efficient linear optimization oracle, allowing the same algorithmic
%     template to be used across different combinatorial structures.

%     \item \textbf{Empirical evaluation.}
%     We evaluate the method on Maximum Coverage, Quadratic Assignment,
%     Max Cut, and Densest $k$-Subgraph, where it achieves strong solution
%     quality while substantially reducing the sequential computation required
%     by existing decomposition-based approaches.
% \end{itemize}
\vspace{-0.3cm}
\section{Problem Definition and Learning Setup}
\label{sec:problem}
\vspace{-0.1cm}
We consider combinatorial optimization (CO) problems over a finite feasible
set $\mathcal{C}\subseteq D^n$, where $D$ is a finite discrete domain. A feasible
solution $\bx\in\mathcal{C}$ assigns a value in $D$ to each of the $n$
decision variables $\bx_1,\ldots,\bx_n$, and is evaluated by an objective
$f:\mathcal{C}\rightarrow\mathbb{R}$. We write the problem as
\begin{align}
    \max f(\bx) \quad \text{s.t.} \quad \bx\in\mathcal{C}.
    \label{eq:main-co}
\end{align}
% This formulation covers, among others, subset-selection problems, discrete assignments, and permutations.

% Many problems considered in this work have Boolean decision variables, $D=\{0,1\}$ and $\mathcal{C}\subseteq\{0,1\}^n$. For example, in Maximum Coverage, $\bx_i=1$ indicates that set $i$ is selected, subject to a cardinality budget, and $f(\bx)$ measures the resulting coverage. 
\rev{
We associate $\mathcal{C}$ with its combinatorial polytope
$\mathcal{P}=\operatorname{conv}(\mathcal{C})$,
the set of all convex combinations of feasible solutions.
Every vertex of $\mathcal{P}$ belongs to $\mathcal{C}$,
but not every point in $\mathcal{C}$ is necessarily a vertex.} Although $\mathcal{P}$ may have exponentially many
constraints or vertices, many important CO problems admit efficient linear
optimization over $\mathcal{P}$. We assume access to an efficient Linear
Maximization Oracle (LMO). Given a query vector $\bq\in\mathbb{R}^n$, an
exact LMO returns a feasible solution
\[
    \bv^\star
    \in
    \argmax_{\bv\in\mathcal{C}}
    \langle \bq,\bv\rangle,
\]\rev{
where linear maximization over $\mathcal{C}$ is equivalent to maximization over $\mathcal{P}=\operatorname{conv}(\mathcal{C})$. Although the set of maximizers over $\mathcal{P}$ may form a face, an optimal vertex always exists and can be selected using a fixed tie-breaking rule.
} An approximate LMO returns a feasible solution that approximately
maximizes the same linear objective. Many important combinatorial polytopes admit efficient LMOs. Examples include the hypersimplex, spanning-tree polytopes, and, more generally, matroid polytopes, and the Birkhoff
polytope for which exact deterministic LMOs can
be implemented using standard combinatorial algorithms. Our framework also
allows approximate LMOs when exact linear optimization is unavailable or
computationally expensive.

\noindent\textbf{Self-Supervised CO and training.}
Given a problem instance $\mathcal{I}$ with input features
$\mathbf{Z}_{\mathcal{I}}$, we use a neural network
$\mathrm{NN}_{\theta}$ to generate $m$ query directions
$\bq_1,\ldots,\bq_m\in\mathbb{R}^n$. \rev{Having these directions, we} apply the LMO \rev{to each of them}
independently and in parallel to obtain feasible solutions
$\bv_1,\ldots,\bv_m\in\mathcal{C}$. Hence, unlike standard neural CO
approaches, the network is not required to directly predict a feasible
solution, and no rounding step is needed to obtain the generated
candidates.

Training is self-supervised: no optimal solutions are provided as labels.
Instead, we evaluate the combinatorial objective on the generated feasible
solutions and construct a differentiable loss of these values. In
particular, we define the interpolation point
$\bx_\theta=\frac{1}{m}\sum_{i=1}^m \bq_i$ and associate it with a distribution \rev{$\mathcal D_{\bx_\theta}$} over
$\bv_1,\ldots,\bv_m$.
\rev{
The distribution assigns weights $\boldsymbol{\alpha}_i(\bx_\theta)$ to the generated feasible solutions,
allowing their objective values to define a differentiable training
signal with respect to the interpolation point $\bx_\theta$ i.e. $F(\bx_\theta) = \sum_{i=1}^m \boldsymbol{\alpha}_i(\bx_\theta)f(\bv_i)=\E_{\bv_i\sim\mathcal{D}_{\bx_\theta}}[f(\bv_i)]$.
}
The resulting \rev{self-supervised} loss provides a gradient that is
backpropagated through the sampled queries to train
$\mathrm{NN}_{\theta}$. At inference time, the same parallel LMO procedure
directly produces feasible discrete solutions, from which the best
candidate can be selected according to the objective. \rev{See \Cref{fig:overall-pipline} for an overview of our pipeline.}
\vspace{-0.3cm}
\section{Method}
\label{sec:method}
\vspace{-0.1cm}
We introduce a parallel alternative to sequential decomposition-based
extensions \rev{in recent works~\citep{karalias2025geometric,nerem2025differentiable,Rafiey2026-FWNCO}. Removing the sequential dependency is a central technical contribution of our work.} The main idea is simple: rather than constructing feasible
vertices one at a time, we generate a collection of query directions in
parallel and solve the corresponding linear maximization problems
simultaneously. The resulting feasible vertices define a local surrogate
extension around the current neural network output. \rev{During training, we solve the simplex Quadratic Program (QP) to identify the positive support, then use implicit differentiation to compute the backward signal through a single KKT solve.}

% Importantly, training
% does not require explicitly computing the convex-combination coefficients as
% we obtain the gradient of the loss directly through implicit
% differentiation.

% \paragraph{Problem setup.} Consider a combinatorial optimization problem with feasible solutions
% $\mathcal{C}\subseteq\mathbb{R}^d$ and objective
% $f:\mathcal{C}\rightarrow\mathbb{R}$. We denote the convex hull of the
% feasible solutions by
% \[
%     \mathcal{P}
%     :=
%     \operatorname{conv}(\mathcal{C}).
% \]
% We assume access to a linear minimization oracle (LMO) over $\mathcal{P}$,
% \[
%     \operatorname{LMO}(q)
%     \in
%     \arg\min_{v\in\mathcal{P}} \langle q,v\rangle.
% \]
% Since the optimum of a linear objective over a polytope is attained at a
% vertex, the LMO returns a feasible combinatorial solution.

% Given an instance $X$, let $x\in\mathbb{R}^d$ denote the continuous
% representation produced by the neural model.
% Our goal is to construct a small collection of feasible vertices around
% $x$ without running a sequential decomposition algorithm.

\noindent\textbf{Representation and parallel query generation.}
Let $\mc I$ an an instance of the CO problem. We first generate $m$ query directions $\bq_1,\ldots,\bq_m\in\mathbb{R}^n$
using a neural network and define $\bx_\theta=\frac{1}{m}\sum_{i=1}^m \bq_i \in\mathbb{R}^n$ to be the continuous representation of the instance $\mc I$ produced by the neural model. (More generally, any differentiable aggregation of the query directions could be used.) \rev{
The interpolation point $\bx_\theta$ serves as a differentiable reference
for the loss construction described below and is not required
to be a feasible solution.} The queries are generated jointly and therefore
do not depend sequentially on one another. There are several possible parameterizations for the query directions.
Importantly, each query $\bq_i$ must have the same dimension as a feasible
solution, i.e., $\bq_i\in\mathbb{R}^n$ when
$\mathcal{C}\subseteq\mathbb{R}^n$.

For example, for graph problems, one implementation is to associate $m$ Gaussian
distributions with each coordinate of the feasible solution. For instance,
when a feasible solution is represented by one coordinate per graph vertex,
the neural network outputs, for every vertex $j\in[n]$, $
    (\mu_{1j},\sigma_{1j}),\ldots,(\mu_{kj},\sigma_{kj}).$
We then sample
\begin{align}
\label{eq:gauss1}
    q_{ij}
    =
    \mu_{ij}+\sigma_{ij}\epsilon_{ij},
    \qquad
    \epsilon_{ij}\sim\mathcal{N}(0,1),
\end{align}
and form the $i$th query as $ \bq_i=(q_{i1},\ldots,q_{in})\in\mathbb{R}^n$, for all $i\in[m]$.
Thus, the network parameterizes $m$ distributions over directions in the
same space as the feasible solutions, and all $m$ queries can be sampled
in parallel.

Alternatively, the network can directly output $(\bq_1,\ldots,\bq_m)=\text{NN}_\theta(\mc{I})$, without introducing a stochastic parameterization. Our framework is
agnostic to the particular parameterization; the important property is
that all $m$ query directions are generated simultaneously and can
therefore be passed to the LMO in parallel. This provides a stochastic mechanism for exploring different regions of
the feasible polytope. 

% Let $\bx_\theta=\frac{1}{m}\sum_{i=1}^m \bq_i \in\mathbb{R}^n$ be the continuous representation of the instance $\mc I$ produced by the neural model. 

\noindent\textbf{Parallel linear maximization and stability.}
For every query $\bq_i$, we use the deterministic LMO to obtain $\bv_i\in \mc C$ which (approximately) maximizes $\langle \bq_i,\bv\rangle$ over $\bv \in \mc C$. These $m$ LMO calls are independent and can therefore be evaluated in
parallel. This is the main computational difference from sequential decomposition
methods. In those methods, the query at iteration $t+1$ depends on the
vertex returned at iteration $t$. Here, all $\bq_i$ are available before
any LMO is evaluated, and all $\bv_i$ can consequently be computed in one
batched parallel stage. We collect the resulting feasible vertices into the matrix
\[
    V
    :=
    \begin{bmatrix}
        \bv_1 & \bv_2 & \cdots & \bv_m
    \end{bmatrix}
    \in\mathbb{R}^{n\times m}.
\]

Although the query directions $\bq_1,\ldots,\bq_m$ vary
continuously with the neural network output, the feasible
solutions selected by the oracle need not change with every
small perturbation. \rev{Each query defines a linear score on the finite feasible set $\mc C$. If one solution scores strictly higher than all others, sufficiently small perturbations preserve its lead, so an LMO returns the same solution.} For example, a top-$k$ oracle returns the
same set as long as the perturbation does not cause a selected
element to fall below an unselected one. Thus, the generated
vertex matrix $V$ can remain fixed even as the queries and
their mean $\bx_\theta$ change.

We assume that the oracle, together with its fixed tie-breaking
rule, has this local stability for almost every collection of
query directions: sufficiently small perturbations leave all
selected vertices unchanged. The exceptional inputs, where an
arbitrarily small perturbation can change a selected vertex,
are assumed to form a set of Lebesgue measure zero. This
property holds for many combinatorial oracles: top-$k$ selection, standard matroid greedy
algorithms such as Kruskal's algorithm, and exact assignment
or matching oracles with fixed tie-breaking.

\vspace{-0.1cm}
\noindent\textbf{The loss over the generated vertices.}
The generated vertices define the restricted polytope $\mathcal{P}_V:=\operatorname{conv}\{\bv_1,\ldots,\bv_m\}.$ \rev{
We use the interpolation point $\bx_\theta$ to assign weights
to the generated feasible candidates $\bv_1,\ldots,\bv_m$
through a simplex-constrained projection of $\bx_\theta$ into $\mathcal{P}_V$.
This construction does not require $\bx_\theta$ to lie
in $\mathcal{P}_V$, allowing us to define a training
surrogate without an exact decomposition of $\bx_\theta$.
} In particular, define
\begin{align}
    \boldsymbol\alpha(\bx_\theta)
    \in
    \argmin_{\boldsymbol\alpha\in\mathbb{R}^m}
    &\quad
    \frac{1}{2}
    \left\|\bx_\theta-V\boldsymbol\alpha\right\|_2^2
    \label{eq:alpha_projection-not-convex}
    \qquad
    \text{s.t.}
    \qquad
    \mathbf{1}^{\top}\balpha=1, \,\,\, \balpha \geq 0.
\end{align}

% \begin{align}
%     \alpha(x)
%     \in
%     \arg\min_{\alpha\in\mathbb{R}^k}
%     &\quad
%     \frac{1}{2}
%     \left\|x-V\alpha\right\|_2^2 + \frac{\lambda}{2}\|\alpha\|^2
%     \label{eq:alpha_projection}\\
%     \text{s.t.}
%     &\quad
%     \mathbf{1}^{\top}\alpha=1,
%     \qquad
%     \alpha\geq 0.
%     \nonumber
% \end{align}

The coefficients $\balpha(\bx_\theta) \in \Delta^m$ define a distribution over the generated feasible solutions. This would naturally induce the expected loss defined as
\begin{align}
\label{eq:extension}
    F_V(\bx_\theta)
    :=
    \sum_{i=1}^m
    \balpha_i(\bx_\theta)f(\bv_i).
\end{align}

To support gradient-based training, we seek a surrogate loss
$F_V$ that is differentiable almost everywhere (a.e.) with
respect to $\bx_\theta$. Next, we establish this and then derive a direct procedure for computing the gradient
from the associated KKT conditions.

% A direct implementation would therefore solve
% \eqref{eq:alpha_projection-not-convex} at every training step and then evaluate
% $F_V(\bx_\theta)$. Although \eqref{eq:alpha_projection-not-convex} is a small convex quadratic
% program, repeatedly solving it introduces an additional optimization
% problem inside the training loop. Our key observation is that this is unnecessary for gradient-based
% learning. Instead of explicitly computing $F_V(\bx_\theta)$, we directly compute its gradient with respect to $\bx_\theta$.

\vspace{-0.1cm}
\noindent\textbf{Computing the gradient directly.} The key observation is that, once the generated vertices $V$ are fixed followed by our oracle stability assumption, the coefficients $\balpha(\bx_\theta)$ are locally differentiable functions of the continuous point $\bx_\theta$. Therefore, to optimize $F_V(\bx_\theta):=\sum_{i=1}^m\balpha_i(\bx_\theta)f(\bv_i)$
it is sufficient to compute the Jacobian $J(\bx_\theta)=\partial \balpha(\bx_\theta)/\partial \bx_\theta.$

The generated vertices $\bv_1,\ldots,\bv_m$ need not be affinely
independent, and therefore the coefficients representing a point in their
convex hull need not be unique. This breaks differentiability. To obtain a unique distribution, we add a
small quadratic regularizer with $\gamma >0$ and define
\begin{equation}
\label{eq:alpha_projection}
    \boldsymbol\alpha(\bx_\theta)
    :=
    \argmin_{\balpha\in \R^m}
    \frac{1}{2}\|\bx_\theta-V\balpha\|_2^2
    +
    \frac{\gamma}{2}\|\balpha\|_2^2
    \qquad
    \text{s.t.}
    \qquad
    \mathbf{1}^{\top}\balpha=1, \,\,\, \balpha \geq 0.
\end{equation}
Since the Hessian with respect to $\balpha$ is
$V^\top V+\gamma I\succ0$ (a positive definite matrix), the objective is strongly convex and
\eqref{eq:alpha_projection} admits a unique solution. Moreover,
$\balpha(\bx_\theta)$ lies on the probability simplex and therefore defines a
distribution over the generated feasible solutions.

To differentiate the coefficients $\balpha(\bx_\theta)$ with respect to
$\bx_\theta$, we first need a set of equations that characterizes the
optimal solution of \eqref{eq:alpha_projection}. For constrained
optimization problems, these are given by the
Karush--Kuhn--Tucker (KKT) optimality conditions, which generalize the
usual condition of setting the gradient of an unconstrained objective to
zero. The constraint $\mathbf{1}^\top\balpha=1$ is associated with a scalar
Lagrange multiplier $\lambda$, while each nonnegativity constraint
$\balpha_i\geq0$ has a multiplier $\boldsymbol \mu_i\geq0$. Since the optimal
coefficients depend on $\bx_\theta$, these multipliers generally depend on
$\bx_\theta$ as well. The stationarity condition for
\eqref{eq:alpha_projection} is $(V^\top V+\gamma I)\balpha
    -V^\top\bx_\theta
    +\lambda\mathbf{1}
    -\boldsymbol \mu
    =0.$
% \begin{equation}
%     (V^\top V+\gamma I)\balpha
%     -V^\top\bx_\theta
%     +\lambda\mathbf{1}
%     -\boldsymbol \mu
%     =0.
% \end{equation}
Together with the original constraints, the KKT conditions additionally
require complementary slackness,
$\boldsymbol \mu_i\balpha_i=0$ for every $i$. This last condition has a simple interpretation. If a coefficient is
strictly positive, $\balpha_i>0$, then its nonnegativity constraint is not
active and therefore $\boldsymbol\mu_i=0$. Let
$S=S(\bx_\theta):=\{i:\alpha_i(\bx_\theta)>0\}$ denote the set of such
coefficients, and let $V_S=[\bv_i]_{i\in S}$. On a region where this
support remains unchanged, coefficients outside $S$ remain zero, while
the positive coefficients satisfy
\begin{equation}
\label{eq:kkt-positive}
    \underbrace{
    \begin{bmatrix}
        V_S^\top V_S+\gamma I & \mathbf{1}\\
        \mathbf{1}^\top & 0
    \end{bmatrix}}_{K_S}
    \begin{bmatrix}
        \balpha_S(\bx_\theta)\\
        \lambda
    \end{bmatrix}
    =
    \begin{bmatrix}
        V_S^\top\bx_\theta\\
        1
    \end{bmatrix}.
\end{equation}
The first row of \eqref{eq:kkt-positive} is the stationarity condition restricted
to the positive coefficients, while the second row enforces
$\mathbf{1}^\top\balpha_S=1$. Thus, once the support $S$ is fixed, the
optimal coefficients are characterized by a linear system whose
right-hand side depends linearly on $\bx_\theta$. This form will allow us
to differentiate $\balpha_S(\bx_\theta)$ implicitly without
differentiating through the optimization procedure itself.

Because $\gamma>0$, $K_S$ is nonsingular. Differentiating
\eqref{eq:kkt-positive} with respect to $\bx_\theta$ gives
\begin{equation}
\label{eq:implicit_jacobian}
    \begin{bmatrix}
        \dfrac{\partial\balpha_S(\bx_\theta)}{\partial \bx_\theta}\\[1mm]
        \dfrac{\partial\lambda}{\partial \bx_\theta}
    \end{bmatrix}
    =
    K_S^{-1}
    \begin{bmatrix}
        V_S^\top\\
        0
    \end{bmatrix}.
\end{equation}
The derivative of the Lagrange multiplier ${\partial\lambda}/{\partial \bx_\theta}$ is only an auxiliary quantity;
we retain the first block of the solution, which gives
$J_S=\partial\balpha_S(\bx_\theta)/\partial \bx_\theta
\in\mathbb{R}^{|S|\times n}$. Let
$f_S=(f(\bv_i))_{i\in S}$ collect the objective values of the vertices
with positive probability. The resulting loss is
$F_V(\bx_\theta)=\balpha(\bx_\theta)^\top f$, and within a region where the support and
generated vertices remain fixed,
\begin{equation}
\label{eq:extension_gradient}
        \nabla_{\bx_\theta} F_V(\bx_\theta)
        =
        J_S^\top f_S.
\end{equation}

Importantly, we do not need to explicitly construct the Jacobian $J_S$.
From \eqref{eq:implicit_jacobian}, $J_S
    =
    \begin{bmatrix} I & 0 \end{bmatrix}
    K_S^{-1}
    \begin{bmatrix}
        V_S^\top\\
        0
    \end{bmatrix}.$ Substituting this expression into
\eqref{eq:extension_gradient} gives
$
    \nabla_{\bx_\theta}F_V(\bx_\theta)
    =
    \begin{bmatrix}V_S & 0\end{bmatrix}
    K_S^{-1}
    \begin{bmatrix}
        f_S\\
        0
    \end{bmatrix}$. Hence, rather than forming $K_S^{-1}$ or $J_S$, we solve the linear
system
\begin{equation}
\label{eq:adjoint_kkt}
    K_S
    \begin{bmatrix}
        \bz\\
        \eta
    \end{bmatrix}
    =
    \begin{bmatrix}
        f_S\\
        0
    \end{bmatrix},
\end{equation}
where $\eta$ is an auxiliary scalar. The extension gradient is then
obtained directly as
\begin{align} \label{eq:dev_F}
    \nabla_{\bx_\theta}F_V(\bx_\theta)=V_S\bz.
\end{align}

\begin{wrapfigure}{r}{0.5\textwidth}
\vspace{-\baselineskip}
\begin{minipage}{0.5\textwidth}
\begin{algorithm}[H]

% \begin{algorithm}[t]
\caption{Gradient Computation}
\label{alg:parallel-extension}
\begin{algorithmic}[1]
\Require Instance $\mc I$, query network $\text{NN}_\theta$, LMO, objective $f$,
number of queries $m$, regularization $\gamma>0$

\State Generate and sample query directions
$\bq_1,\ldots,\bq_m \gets \text{NN}_\theta(\mc I)$
\State Set the interpolation point
$\bx_\theta \gets \frac{1}{m}\sum_{i=1}^m \bq_i$

\State \textbf{in parallel:}
$\bv_i \gets \operatorname{LMO}(\bq_i)$ for $i=1,\ldots,m$
\State $V \gets [\bv_1,\ldots,\bv_m]$

\State Compute: $\balpha
    \gets
    \argmin_{\balpha\geq 0}
    \frac{1}{2}\|\bx_\theta-V\balpha\|_2^2
    +
    \frac{\gamma}{2}\|\balpha\|_2^2
    \quad
    \text{s.t.}\quad
    \mathbf{1}^\top\balpha=1$
% \[
%     \balpha
%     \gets
%     \argmin_{\balpha\geq 0}
%     \frac{1}{2}\|\bx-V\balpha\|_2^2
%     +
%     \frac{\gamma}{2}\|\balpha\|_2^2
%     \quad
%     \text{s.t.}\quad
%     \mathbf{1}^\top\balpha=1
% \]

\State $S \gets \{i:\balpha_i>0\}$, $V_S \gets [\bv_i]_{i\in S}$, and $f_S \gets (f(\bv_i))_{i\in S}$

\State Form $K_S
    \gets
    \begin{bmatrix}
        V_S^\top V_S+\gamma I & \mathbf{1}\\
        \mathbf{1}^\top & 0
    \end{bmatrix}$
% \[
%     K_S
%     \gets
%     \begin{bmatrix}
%         V_S^\top V_S+\gamma I & \mathbf{1}\\
%         \mathbf{1}^\top & 0
%     \end{bmatrix}
% \]

\State Solve for $\bz$: $K_S
    \begin{bmatrix}
        \bz\\
        \eta
    \end{bmatrix}
    =
    \begin{bmatrix}
        f_S\\
        0
    \end{bmatrix}$
% \[
%     K_S
%     \begin{bmatrix}
%         \bz\\
%         \eta
%     \end{bmatrix}
%     =
%     \begin{bmatrix}
%         f_S\\
%         0
%     \end{bmatrix}
% \]

\State $\nabla_{\bx_\theta} F_V(\bx_\theta) \gets V_S \bz$

\State \Return $\nabla_{\bx_\theta} F_V(\bx_\theta)$
% \end{algorithmic}
% \end{algorithm}

\end{algorithmic}
\end{algorithm}
\end{minipage}
\vspace{-2.1\baselineskip}
\end{wrapfigure}

Therefore, after determining the positive support, computing the backward
signal requires only a linear solve with the KKT matrix; neither the
Jacobian nor its matrix inverse is formed explicitly.

\rev{
To propagate the gradient to the network parameters,
consider the training objective
$L(\theta)\coloneqq F_{V(\theta)}(\bx_\theta)$, where
$\bx_\theta=\frac{1}{m}\sum_{i=1}^{m}\bq_i(\theta)$.
When sufficiently small perturbations of the queries leave
the LMO outputs unchanged, $V(\theta)$ remains locally fixed.
The dependence of $L$ on $\theta$ is then through the
interpolation point $\bx_\theta$, and the chain rule gives
\begin{equation} \label{eq:chain}
\hspace{-0.3cm}\nabla_\theta L(\theta)=\frac{1}{m}\sum_{i=1}^{m}
\left(\frac{\partial \bq_i}{\partial\theta}\right)^\top
\nabla_{\bx_\theta}F_V(\bx_\theta).
\end{equation}
Combining this expression with \eqref{eq:dev_F} gives the gradient used to update the network parameters without differentiating through the discrete LMO.
}

Note that, for any fixed support $S$, the KKT system shows that
$\balpha_S(\bx_\theta)$ is an affine function of $\bx_\theta$.
As $\bx_\theta$ varies, the support may change, giving different affine
pieces. Hence $\balpha(\bx_\theta)$ is piecewise affine and
$F_V(\bx_\theta)$ is differentiable almost everywhere.
\rev{Consequently, the training objective $L(\theta)=F_{V(\theta)}(\bx_\theta)$ is also differentiable almost everywhere with respect to $\theta$, with its gradient given by \eqref{eq:chain}.} The procedure is given in \Cref{alg:parallel-extension}.

\begin{proposition}[A.e. differentiability and gradient correctness]
\label{prop:aediff}
Let $\mathcal P=\operatorname{conv}(\mathcal C)$ admit a deterministic LMO with fixed tie-breaking, and let $\gamma>0$. Assume the network queries $\bq_i(\theta)$ are locally Lipschitz and the generated vertices $V(\theta)$ are locally constant for a.e.\ $\theta$. Let $\bx= \frac{1}{m}\sum_{i=1}^m\bq_i(\theta)$. For fixed $V$, \eqref{eq:alpha_projection} admits unique coefficients $\balpha(\bx)\geq0$ with $\mathbf1^\top\balpha(\bx)=1$, and $\balpha(\bx)$ is piecewise affine and hence differentiable a.e.\ in $\bx$. Consequently, $L(\theta)=F_{V(\theta)}(\bx_\theta)$ is differentiable a.e.\ in $\theta$, and \Cref{alg:parallel-extension}, followed by backpropagation through $\bx_\theta$, computes the exact gradient a.e.
\end{proposition}

% \begin{proposition}[A.e. differentiability and gradient correctness]
% \label{prop:aediff}
% Let $\mathcal P=\operatorname{conv}(\mathcal C)$ admit
% a deterministic LMO, and let $\gamma>0$.
% Assume the network queries $\bq_i(\theta)$ are locally
% Lipschitz and the generated vertices $V(\theta)$ are
% locally constant for a.e.\ $\theta$.
% Algorithm~1 computes unique coefficients
% $\balpha\ge0$ with $\mathbf1^\top\balpha=1$ that,
% for fixed $V$, are differentiable a.e.\ in $\bx$.
% Moreover, the loss $F_{V(\theta)}(\bx_\theta)$,
% where $\bx_\theta=m^{-1}\sum_i\bq_i(\theta)$,
% is differentiable a.e.\ in $\theta$.
% Algorithm~1, followed by backpropagation through
% $\bx_\theta$, computes its gradient a.e.
% \end{proposition}

\vspace{-0.1cm}
\noindent\textbf{Inference.} At inference time, the network generates $m$ query directions, which are
passed independently through the LMO to obtain feasible candidates
$\bv_1,\ldots,\bv_k\in\mathcal{C}$. Since feasibility is enforced by the
oracle, no rounding step is required. We evaluate the combinatorial
objective on the generated candidates and return the best one. See \cref{fig:overall-pipline} for the overall pipeline in inference and training.
\rev{
\begin{myremark}
Geometric decomposition methods such as
\citet{karalias2025geometric, nerem2025differentiable} require the point
to be decomposed to lie in the feasible polytope
$\mathcal{P}$. Enforcing this condition may require
problem-specific transformations or computationally
demanding projections, particularly for matroid base
and spanning-tree polytopes, which have exponentially
many defining constraints.
Our approach, similar to \cite{Rafiey2026-FWNCO}, imposes no such restriction on the interpolation point
$\bx_\theta$. The regularized QP is well-defined
for any $\bx_\theta\in\mathbb{R}^n$ and assigns a
probability distribution over the generated feasible
candidates without requiring projection onto
the full polytope $\mathcal{P}$.
\end{myremark}
}

\begin{myremark}
The constraint $\balpha\geq0$ is required only if the coefficients are to
define a probability distribution over the generated vertices. If it is
removed while retaining $\mathbf{1}^\top\balpha=1$, the coefficients instead
define an affine combination of the vertices, giving a regularized affine
interpolation over their affine hull. In this case there is no active set:
the KKT matrix is fixed for a given $V$, and the gradient can be
computed directly from a linear KKT solve \textbf{without} first solving for
$\balpha(\bx_\theta)$ \eqref{eq:alpha_projection}. Thus, the nonnegative formulation provides a
probabilistic interpretation, while the equality-only formulation yields a
simpler affine extension and backward computation. This makes a speed difference in training, but is irrelevant to inference. 
\end{myremark}

% \noindent\textbf{Overall procedure.} The complete procedure has three stages. First, the neural network
% generates $m$ query directions simultaneously. Second, we perform $m$
% independent LMO calls to obtain $m$ feasible vertices in parallel.
% Finally, we evaluate the combinatorial objective on these vertices and
% use \eqref{eq:extension_gradient} to propagate a gradient through the
% continuous output $\bx_\theta$. See \cref{fig:overall-pipline}.

% \begin{proof}[Proof sketch]
% To Do.
% \end{proof}
% \vspace{-0.3cm}
\section{Experiments}
% \vspace{-0.1cm}
\label{sec:experiment}

We evaluate our framework on Maximum Coverage (MC), Quadratic Assignment (QAP), MaxCut, and Densest $k$-Subgraph (DKS). Our framework applies unchanged across problems; only the LMO changes. The network need only produce $m$ query vectors and support backpropagation through them, allowing architectures and data pipelines tailored to each problem. Our implementations scale from small assignment problems to graphs with millions of vertices, achieve SOTA solution quality in several comparisons, and lie on the quality--time Pareto frontier across all four problems. We exploit GPU parallelism to generate queries in one network forward pass and obtain feasible solutions through independent, parallel LMO calls. Compared with sequential decomposition or iterative neural improvement, this reduces sequential computation and repeated network evaluations during training and inference. This efficiency enables fast large-scale solution generation and opens opportunities to learn better representations from larger, more diverse training instances. Code is available at \href{https://github.com/frankx2023/Parallel-NCO}{Link}.

% We evaluate our framework on four CO problems: Maximum Coverage (MC), Quadratic Assignment (QAP), MaxCut, and Densest $k$-Subgraph (DKS). One advantage of our framework is that it only requires the neural network to produce $m$ query vectors and support backpropagation through them. This flexibility allows us to design distinct neural architectures and data-processing pipelines tailored to each problem's structure and scale. With these designs, our method scales from small assignment problems to graphs with millions of vertices, achieves SOTA solution quality in several comparisons, and lies on the quality--time Pareto frontier across all four problems.

% Our method exploits GPU parallelism by generating multiple query vectors
% in a single neural network forward pass and producing feasible solutions
% through independent, parallel LMO calls. Compared with sequential
% decomposition or iterative neural improvement, this design reduces
% sequential computation and repeated network evaluations during both
% training and inference. The resulting efficiency enables fast solution
% generation for large-scale problems and creates opportunities to train
% on larger and more diverse instances, improving the scope for learning
% effective problem representations. The code is available at \href{https://anonymous.4open.science/r/Parallel_NCO-1DEC/README.md}{Link}.

\vspace{-0.3cm}
\subsection{Maximum Cut}
\label{sec:exp-maxcut}
\vspace{-0.1cm}

\begin{wrapfigure}{r}{0.45\textwidth}
\vspace{-\baselineskip}
\begin{minipage}{0.45\textwidth}

% \begin{wrapfigure}[26]{r}{0.5\textwidth}
% \begin{figure}[t]
  \centering
  %%%%%%%%%
%\includegraphics[width=0.9\linewidth]{figures/maxcut-panel.pdf}
  % Figure used for submission
  %%%%%%%%%%%%
  \includegraphics[width=1\linewidth]{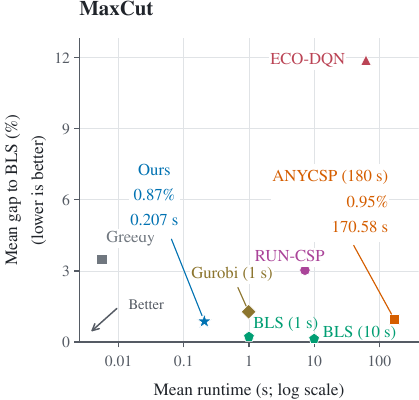}
  \caption{MaxCut on 35 Gset graphs: mean
    gap to the BLS reference versus mean run-time (\emph{log scale}), with equal weight per graph. \rev{BLS is additionally evaluated at fixed time budgets of 1 and 10 seconds.}
    Lower and left are better.}
  \label{fig:maxcut-quality-runtime}
% \end{figure}
% \end{wrapfigure}
\end{minipage}
\vspace{-\baselineskip}
\end{wrapfigure}

Let $G=(V,E)$ be an undirected graph with $n$ vertices and edge weights
$w_{uv}$. MaxCut seeks a partition of the vertices into
two parts that maximizes the total weight of crossing edges. Representing the partition by $\mathbf s\in\{-1,1\}^n$, the
problem is $\max_{\mathbf s\in\{-1,1\}^n}f_{\text{cut}}(\mathbf s)$,
where $f_{\text{cut}}(\mathbf s)
=\sum_{\{u,v\}\in E}w_{uv}(1-\mathbf s_u\mathbf s_v)/2$ and each edge is counted
once. Note that, MaxCut has no coupling constraint between variables: every sign vector is feasible, so
$\mathcal{C}=\{-1,1\}^n$.

% A vector $\mathbf s\in\{-1,1\}^n$ assigns each vertex to one of two
% parts. The value of the resulting cut is $ f_{\text{cut}}(\mathbf s)
%  =\sum_{\{u,v\}\in E}w_{uv}\frac{1-\mathbf s_u\mathbf s_v}{2}.$ The MaxCut problem is $\max_{\mathbf s\in\{-1,1\}^n}f_{\text{cut}}(\mathbf s)$,
% where each undirected edge is counted once. Note that, MaxCut has no coupling constraint between variables: every sign vector is feasible, so
% $\mathcal{C}=\{-1,1\}^n$.

\vspace{-0.1cm}
\noindent\textbf{Neural network, oracle, and training.}
A width-32 residual message-passing network with 512 layers processes
BFS parity features and random node features. A learned mapping combines
the resulting node embeddings with fixed Sobol vectors to produce
$\bq_1,\ldots,\bq_m\in\mathbb{R}^n$.
We use the node-assignment hypercube
$\mathcal{P}_{\text{cut}}=\operatorname{conv}(\{-1,1\}^n)=[-1,1]^n$.
Its LMO acts independently on each coordinate:
$\bv_{i,u}=1$ if $\bq_{i,u}\geq0$, and $\bv_{i,u}=-1$ otherwise.
Each returned vertex therefore defines a cut.

We supply normalized negative cut values to Algorithm~\ref{alg:parallel-extension} and propagate
the gradient through the mean query to the learned mapping
and message-passing network. Training uses randomly generated graphs
of mixed families and sizes, with $m=16$.
Following ANYCSP~\citep{tonshoff2022one}, we evaluate on 35 unweighted
Gset instances with 800--10000 vertices. At inference, we use $m=1000$
and return the assignment attaining $\max_j f_{\text{cut}}(\bv_j)$.

\noindent\textbf{Results.}
Figure~\ref{fig:maxcut-quality-runtime} summarizes the quality--runtime
trade-off across the 35 Gset graphs. Points represent results at the evaluated budgets, not time to a common target quality. SDP is omitted because an overall mean over all 35
  graphs is unavailable. \rev{Among the compared neural methods,} our method achieves the lowest
reported overall mean gap to the BLS~\citep{benlic2013breakout} reference, $0.87\%$, with a mean
runtime of $0.207$ seconds per graph. This improves over ANYCSP's mean gap of $0.95\%$ while being more than
$800\times$ faster, and also improves over the $1$-second Gurobi baseline
both in solution quality and runtime. Greedy is faster but produces substantially worse solutions, with a mean gap of $3.48\%$. The full results, including the breakdown by graph size, are provided in Table~\ref{tab:completed-cut} in
Appendix~\ref{sec:appendix-maxcut}. The quality advantage is particularly pronounced on the largest instances Gset3000+:
our method achieves a mean gap of $0.58\%$, compared with $1.35\%$
for Gurobi and $1.61\%$ for ANYCSP, while requiring only $0.206$
seconds per graph on average. \rev{Recall that all gaps are measured relative to the published BLS reference cuts.
For completeness, we additionally evaluate budgeted BLS as a classical
non-neural metaheuristic baseline specialized for MaxCut. It achieves
lower mean gaps of $0.2245\%$ and $0.1371\%$ at fixed budgets of $1$
and $10$ seconds, respectively. These budgets are approximately
$4.8\times$ and $48.3\times$ our method's mean runtime of $0.207$
seconds.} Further experimental details are provided in
Appendix~\ref{sec:appendix-maxcut}, this includes ablation on $m$ and raw numbers.

\vspace{-0.3cm}
\subsection{Quadratic Assignment Problem}
\label{sec:exp-qap}
\vspace{-0.1cm}

QAP seeks a bijection between $n$ facilities and $n$ locations
that minimizes the total flow-weighted distance, given the
flows between facilities and distances between locations.  Such a
bijection can be represented by a permutation matrix. Given flow and distance matrices
$\mathbf A,\mathbf B\in\mathbb R^{n\times n}$, respectively,
the problem is $\min_{\mathbf P\in\Pi_n}f_{\text{qap}}(\mathbf P)$,
where $f_{\text{qap}}(\mathbf P)
=\mathrm{trace}(\mathbf A^\top\mathbf P\mathbf B\mathbf P^\top)$
and $\Pi_n$ denotes the set of $n\times n$ permutation matrices.
Thus, $\mathcal C=\Pi_n$.

% There are $n$ facilities and $n$ locations. Distances between locations are given by a matrix $\mathbf{B}\in\mathbb{R}^{n\times n}$, and flows between facilities by a matrix $\mathbf{A}\in\mathbb{R}^{n\times n}$. The goal is to assign each facility to a distinct location so as to minimize the total flow-weighted distance i.e., finding a permutation matrix. In QAP, the set of feasible solutions is $\mc C=\Pi_n$ where $\Pi_n$ denotes the set of $n\times n$ permutation matrices. The objective is $\min_{\mathbf{P} \in \Pi_{n}} f_{\text{qap}}(\mathbf{P})$ where $f_{\text{qap}}(\mathbf{P})=\mathrm{trace}(\mathbf{A} \mathbf{P} \mathbf{B} \mathbf{P}^\top)$ for all $\mathbf{P} \in \Pi_{n}$.

\vspace{-0.1cm}
\noindent\textbf{Neural network, oracle, and training.}
GraphSAGE-based encoders process the flow and distance matrices and
predict Gaussian parameters for $m$ assignment-score matrices
$\mathbf Q_j\in\mathbb{R}^{n\times n}$. The feasible polytope is the Birkhoff polytope whose vertices are permutations: $\mathcal{B}_n=\operatorname{conv}(\Pi_n)
 =\{\mathbf P\in\mathbb{R}_{\geq0}^{n\times n}:
\mathbf P\mathbf{1}=\mathbf{1}, \mathbf P^{\top}\mathbf{1}=\mathbf{1}\}.$
% \[
%  \mathcal{B}_n=\operatorname{conv}(\Pi_n)
%  =\{\mathbf P\in\mathbb{R}_{\geq0}^{n\times n}:
%  P\mathbf{1}=\mathbf{1},\ P^{\top}\mathbf{1}=\mathbf{1}\}.
% \]
 We use a greedy weighted-matching algorithm as an
approximate LMO. It selects high-scoring assignments while
enforcing one assignment per row and column, producing a permutation
$\mathbf P_j$ for each query. As in Algorithm \ref{alg:parallel-extension}, the KKT gradient at the mean assignment-score matrix is then calculated and backpropagated through Gaussian sampling and the encoders.
We generate synthetic QAP instances following the procedure of
SAWT~\citep{tan2024learning} and train our model on synthetic QAP32 instances. We evaluate on 14 QAPLIB classes with $12 \leq n \leq 64$.
Inference returns the permutation with cost $\min_j f_{\text{qap}}(\mathbf P_j)$.

\begin{wrapfigure}{r}{0.45\textwidth}
\vspace{-\baselineskip}
\begin{minipage}{0.45\textwidth}
% \begin{wrapfigure}[22]{r}{0.5\textwidth}
  % \centering
  %%%%%% %%%%%% 
  %\includegraphics[width=0.9\linewidth]{figures/qap-narrow.pdf} 
  % the figure used for submission
  %%%%%% %%%%%% 
  \includegraphics[width=1\linewidth]{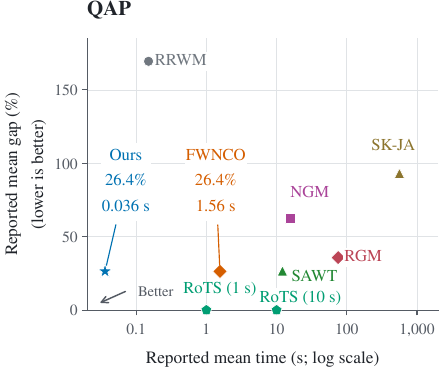}
  \caption{QAPLIB: mean optimality gap
    across 14 classes versus reported mean inference time (\emph{log scale}), with equal weight per class. Lower and left are better.
% QAPLIB solution quality and reported inference time.
%   All seven methods with numerical results in the full table are shown.
%   The vertical axis is the unweighted mean of the 14 displayed per-class
%   optimality gaps, recomputed from the rounded class values in
%   Table~\ref{tab:qap-mean-time}; the horizontal axis shows reported mean
%   inference time on a logarithmic scale. Lower values are better on both
%   axes.
    }
  \label{fig:qap-quality-runtime}
% \end{wrapfigure}
\end{minipage}
\vspace{-\baselineskip}
\end{wrapfigure}

\noindent\textbf{Results.} Figure~\ref{fig:qap-quality-runtime} compares seven methods on QAPLIB using the mean optimality gap across 14 classes and mean inference time. Our method and FWNCO~\citep{Rafiey2026-FWNCO} both achieve a mean class gap of $26.4\%$, computed from the rounded per-class values in Appendix Table~\ref{tab:qap-mean-time}. Our method has the lowest reported inference time at $0.036$ seconds, compared with $1.56$ seconds for FWNCO, a roughly $43\times$ difference. Baseline runtimes are taken from prior publications, while ours was measured locally on different hardware, so the comparison is not fully controlled. However, our method would remain faster than all baselines even if their reported runtimes were reduced tenfold. It also achieves the lowest gap \rev{among the compared neural methods} on five classes (\texttt{rou}, \texttt{scr}, \texttt{ste}, \texttt{tho}, and \texttt{wil}) and the second-best gap on several others, combining competitive solution quality with low reported inference time. \rev{We additionally compare against RoTS~\citep{Taillard1991}, a classical
metaheuristic tailored to QAP, which achieves lower mean class gaps of
$0.0439\%$ and $0.0109\%$ at $1$ and $10$ seconds, respectively.
Among the compared
neural methods, our method matches the lowest reported mean gap (tied with FWNCO at $26.4\%$ after rounding) and has the
lowest reported runtime.} Full results and an ablation on $m$ are provided in Appendix~\ref{sec:appendix-qap}.

\vspace{-0.3cm}
\subsection{Maximum Coverage}

Let $G=(L\cup R,E)$ be a bipartite graph with $L=[n]$
and nonnegative weights $w_u$ on vertices $u\in R$.
MC seeks $k$ vertices from $L$ that maximize
the total weight of their neighbors in $R$.
For a selection vector $\bx\in\{0,1\}^n$, the objective is
$f_{\text{cov}}(\bx)=\sum_{u\in R}w_u
\mathbf1[\sum_{i:(i,u)\in E}\bx_i\geq1]$.
The problem is $\max_{\bx\in\mathcal C}f_{\text{cov}}(\bx)$,
where $\mathcal C=\{\bx\in\{0,1\}^n:\mathbf1^\top\bx=k\}$.

% Let $\mathcal{U}$ be a ground set of elements with weights $w_u \ge 0$ for $u\in\mathcal{U}$, and let $\{S_1,\dots,S_n\}$ be subsets $S_i \subseteq \mathcal{U}$. The value of $\bx\in\{0,1\}^n$ is the total weight of covered elements,
% $
%   f_{\text{cov}}(\bx)
%   \;=\;
%   \sum_{u \in \mathcal{U}}
%   w_u \,\mathbf{1}\!\left[\exists\, i \in [n]\ \text{with}\ u \in S_i \ \text{and}\ \bx(i) = 1\right].
% $
% The MC problem under cardinality constraint is $\max_{\bx \in \{0,1\}^n} f_{\text{cov}}(\bx)$ subject to $\sum_{i=1}^n \bx(i) = k$, where $k$ denotes the prescribed cardinality. For this problem the feasible set is $\mc C=\{\bv\in\{0,1\}^n:\mathbf{1}^{\top}\bv=k\}$.

\noindent\textbf{Neural network, oracle, and training.}
% We represent each bipartite graohinstance as a bipartite graph connecting sets to the elements they contain. 
A GraphSAGE \citep{hamilton2017inductive} network predicts $\bq_1,\ldots,\bq_m\in\mathbb{R}^n$ as explained in \eqref{eq:gauss1}.
The feasible polytope is the hypersimplex
$\mathcal{P}_{n,k}=\{\bx\in[0,1]^n:\mathbf{1}^{\top}\bx=k\}$.
Its LMO selects the $k$ largest entries of each query, yielding
the indicator vector $\bv_j$ of a feasible collection of sets. Following
GeoNCO~\citep{karalias2025geometric}, we train on synthetic random
instances and evaluate on Random500, Random1000, Rail, and Twitch with no test time optimization. At inference time, we evaluate all candidates generated
in parallel and return
$\arg\max_j f_{\mathrm{cov}}(\bv_j)$.
We use $m=50$ parallel queries (see \ref{app:mc-query-budget} for an ablation on this).

\begin{figure}[t]
  \centering
  \includegraphics[width=\linewidth]{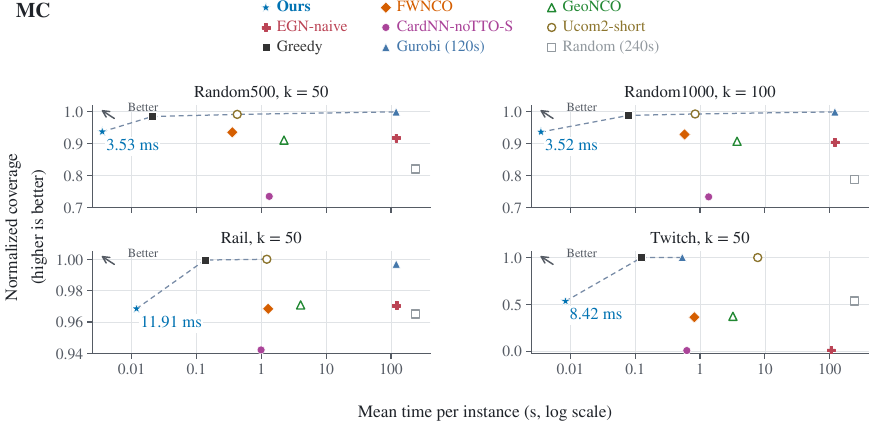}
  \caption{MC at the larger evaluated cardinality bound for
each dataset. Coverage is normalized by the best mean in each
panel; dashed lines show the quality--runtime Pareto frontier.}
  \label{fig:MC-plot}
\end{figure}

\noindent\textbf{Results.}
% Table~\ref{tab:max-coverage-results} reports normalized coverage andinference time. 
Figure~\ref{fig:MC-plot} shows normalized coverage and inference
time for the larger of the two evaluated cardinality bounds
$k$ for each dataset. Full results for both values of $k$
are reported in Appendix Table~\ref{tab:max-coverage-results}. Our method has the lowest inference time in all eight
settings, four of which are shown in th figure, requiring only $0.0035$--$0.0119$ seconds per instance.
Despite this substantially lower runtime, its solution quality remains
competitive with the strongest neural baselines. In particular, it obtains
the highest neural coverage on Random1000 with $k=100$ and on both Twitch settings, while matching the best neural result in several other cases. Because it is the fastest method in every setting, our method lies on the quality--runtime Pareto frontier throughout the benchmark.

% MC is a particularly strong setting for the classical Greedy baseline: its objective is monotone submodular under a cardinality constraint, for which Greedy enjoys the classical $1-1/e$ approximation guarantee. We therefore additionally evaluate on GreedyTrap instances designed to expose cases where the greedy selection rule is less effective. As shown in Table~\ref{tab:mc-greedytrap-large}, our method improves coverage over Greedy by approximately $21$--$30\%$ across these instances. For three of the four settings it is also faster, with the largest case giving an approximately $8.5\times$ speedup.

% \begin{wrapfigure}{r}{0.5\textwidth}
% \vspace{-0.3cm}
% \vspace{-\baselineskip}
% \begin{minipage}{0.5\textwidth}
% \begin{figure}[H]
% % \begin{wrapfigure}{r}{0.5\textwidth}
%   % \vspace*{-\intextsep}
%   \centering
%   %%%%%%%%%
%   % \includegraphics[width=\linewidth]{figures/dks-panel.pdf}
%   % Submission figure
%   %%%%%%%%%
%   \includegraphics[width=\linewidth]{figures/dks_rebuttal_colored.pdf}
%   \caption{DKS on 12 SNAP graphs (k = 100): mean normalized objective versus mean run-time (\emph{log scale}), with equal weight per graph. Higher and left are better}
%   \label{fig:dks-quality-runtime}
% \end{figure}
% % \end{wrapfigure}
% \end{minipage}
% \vspace{-\baselineskip}
% \vspace{-0.8cm}
% \end{wrapfigure}

\begin{wraptable}[13]{r}{0.45\textwidth}
 \vspace*{-\intextsep}
\centering
\scriptsize
\setlength{\tabcolsep}{2pt}
\renewcommand{\arraystretch}{0.9}
\caption{MC normalized coverage and runtime
in seconds. Coverage is normalized by
best mean in each row.}
\label{tab:mc-greedytrap-large}

\begin{tabular}{@{}lrrrrr@{}}
\toprule
& & \multicolumn{2}{c}{\textbf{Greedy}}
  & \multicolumn{2}{c}{\textbf{Ours}} \\
\cmidrule(lr){3-4}\cmidrule(lr){5-6}
Dataset & $k$ & Cov.$\uparrow$ & Time$\downarrow$
             & Cov.$\uparrow$ & Time$\downarrow$ \\
             \midrule
GreedyTrap500
 & 10   & 0.77 & 0.00198   & 1.00 & 0.00362 \\
 & 50   & 0.82 & 0.00710   & 1.00 & 0.00352 \\
GreedyTrap1000
 & 20   & 0.78 & 0.00823   & 1.00 & 0.00382 \\
 & 100  & 0.82 & 0.02950   & 1.00 & 0.00346 \\
\midrule
Random10000
 & 200  & 1.00 & 1.94177   & 0.97 & 0.00633 \\
 & 1000 & 1.00 & 8.10018   & 0.95 & 0.00619 \\
Random50000
 & 1000 & 1.00 & 51.06446  & 0.95 & 0.02492 \\
 & 5000 & 1.00 & 252.55409 & 0.95 & 0.02515 \\
\bottomrule
\end{tabular}
\end{wraptable}

\label{sec:exp-mc}
\vspace{-0.1cm}

Greedy is a strong baseline for MC: its monotone submodular objective gives Greedy a $1-1/e$ approximation guarantee under a cardinality constraint. We evaluate on GreedyTrap instances designed to expose limitations of greedy selection. Our method achieves $21$--$30\%$ higher coverage than Greedy and is faster in three of four settings, reaching an $8.5\times$ speedup on the largest case. We also test scalability on the much larger Random10000 and Random50000 instances. Our method retains $94.8$--$96.6\%$ of Greedy's coverage while using only $0.01$--$0.33\%$ of its runtime, demonstrating high coverage at substantially lower inference cost; see Table~\ref{tab:mc-greedytrap-large}.

% We further evaluate scalability on Random10000 and Random50000.
% Across these larger instances, our method achieves $94.8$--$96.6\%$ of
% Greedy's coverage while using only $0.01$--$0.33\%$ of its runtime.
% These results illustrate the main advantage of the parallel formulation:
% it preserves much of the solution quality of a strong problem-specific
% baseline while substantially reducing inference time as the instance size
% grows.

\subsection{Densest $k$-Subgraph}
\label{sec:exp-dks}

Let $G=(V,E)$ be a simple undirected graph with $n$ vertices
and adjacency matrix $\mathbf A$. DKS
seeks $k$ vertices that induce the largest number of edges.
For a selection vector $\bv\in\{0,1\}^n$, the objective is
$f_{\mathrm{dks}}(\bv)=\sum_{\{u,w\}\in E}\bv_u\bv_w
=\frac12\bv^\top\mathbf A\bv$.
The problem is $\max_{\bv\in\mathcal C}f_{\text{dks}}(\bv)$,
where $\mathcal C=\{\bv\in\{0,1\}^n:\mathbf1^\top\bv=k\}$.

% Let $G=(V,E)$ be a simple undirected graph with $n$ vertices and
% adjacency matrix $\mathbf A$. A vector $\bv\in\{0,1\}^n$ indicates the selected
% vertices, and the number of edges they induce is $ f_{\text{dks}}(\bv)
%  =\sum_{\{u,w\}\in E}\bv_u\bv_w
%  =\frac{1}{2}\bv^{\top}\mathbf A\bv.$ The DKS problem is $\max_{\bv\in\{0,1\}^n}f_{\text{dks}}(\bv)$ subject to
% $\mathbf{1}^{\top}\bv=k$, where $k$ is the prescribed subgraph size. For this problem the feasible set is any subgraph of size $k$; $\mc C=\{\bv\in\{0,1\}^n:\mathbf{1}^{\top}\bv=k\}$.

\vspace{-0.1cm}
\noindent\textbf{Neural network, oracle, and training.}
To scale to million-vertex graphs, we compute compact node features from sampled neighborhoods of the input graph. A shared MLP with hidden widths 32 and 16 combines learned and structural scores to generate full-graph queries $\bq_j\in\mathbb{R}^n$. As in MC, the LMO operates over the hypersimplex $\mathcal{P}_{n,k}$ and selects the $k$ highest-scoring vertices for each query. Thus, the neural network processes only a small portion of the graph, while the LMO produces feasible $k$-vertex subsets over the full graph. In Algorithm~\ref{alg:parallel-extension}, we use negative induced-edge density as the $k$-vertex subsets cost and backpropagate through the mean query to the shared MLP. We train on random 500-vertex graphs with $k\in{20,50,100,200}$ and $m=32$, and evaluate at $k=100$ on the 12 SNAP networks~\citep{leskovec2016snap} used by FW-DL~\citep{lu2025densest} and EP-Prox~\citep{liu2026scalable}. At inference, we use $m=100$ or $m=1000$ and return the subset maximizing $f_{\text{dks}}(\bv_j)$. Queries are processed in chunks of 16 to control memory usage while retaining the full graph.

% To handle million-vertex graphs, we compute compact node features
% relative to sampled neighborhoods on the sparse graph. A shared MLP
% with hidden widths 32 and 16 combines learned scores with structural
% scores to generate full-graph queries $\bq_j\in\mathbb{R}^n$.
% The feasible polytope is the same hypersimplex $\mathcal{P}_{n,k}$ as
% in MC, and its LMO selects the $k$ highest-scoring vertices
% for each query. Such design allows the neural network to process only a small fraction of the graph while the LMO produces a feasible $k$-vertex subgraph on the full graph, demonstrating the scalability of our framework. We use negative induced-edge density as the vertex cost in Algorithm~\ref{alg:parallel-extension}
% and backpropagate its gradient through the mean query to the shared MLP.
% Training uses randomly generated 500-vertex graphs with
% $k\in\{20,50,100,200\}$ and $m=32$.
% We test at $k=100$ on the 12 SNAP networks~\citep{leskovec2016snap}
% used by FW-DL~\citep{lu2025densest} and
% EP-Prox~\citep{liu2026scalable}. At inference, we use $m=100$ or $m=1000$
% and return the subset attaining $\max_j f_{\text{dks}}(\bv_j)$.
% Queries are processed in chunks of 16 to limit memory usage while
% retaining the complete graph.

\begin{wrapfigure}{r}{0.5\textwidth}
\vspace{-0.3cm}
\vspace{-\baselineskip}
\begin{minipage}{0.5\textwidth}
\begin{figure}[H]
% \begin{wrapfigure}{r}{0.5\textwidth}
  % \vspace*{-\intextsep}
  \centering
  %%%%%%%%%
  % \includegraphics[width=\linewidth]{figures/dks-panel.pdf}
  % Submission figure
  %%%%%%%%%
  \includegraphics[width=\linewidth]{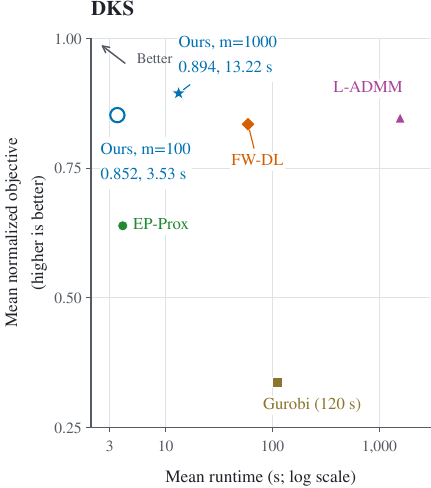}
  \caption{DKS on 12 SNAP graphs (k = 100): mean normalized objective versus mean run-time (\emph{log scale}), with equal weight per graph. Higher and left are better}
  \label{fig:dks-quality-runtime}
\end{figure}
% \end{wrapfigure}
\end{minipage}
\vspace{-\baselineskip}
\vspace{-0.8cm}
\end{wrapfigure}

\vspace{-0.1cm}
\noindent\textbf{Results.}
% Table~\ref{tab:dks-completed} reports normalized objective value and runtime on the 12 SNAP graphs. 
Figure~\ref{fig:dks-quality-runtime} summarizes solution quality and runtime on the 12 SNAP graphs. With only $m=100$ queries, our method achieves a mean normalized objective of $0.85$, matching L-ADMM for the second-best solution quality, while having the lowest mean runtime at $3.53$ seconds. This is about $17\times$ faster than FW-DL and over $400\times$ faster than L-ADMM. Increasing the budget to $m=1000$ improves the mean objective to $0.89$, the best among all compared methods, with a mean runtime of $13.22$ seconds. This remains about $4.5\times$ faster than FW-DL and over two orders of magnitude faster than L-ADMM. It also achieves the best or tied-best objective on several individual graphs, including web-Stanford, soc-Pokec, ca-HepTh, com-DBLP, and soc-LiveJournal1. Overall, $m$ provides a clear quality--runtime trade-off: $m=100$ favors speed, while $m=1000$ improves solution quality at modest additional cost. Full results appear in Table~\ref{tab:dks-completed}, with additional results and ablations in Appendix~\ref{sec:appendix-dks}.

\section{Conclusion}
\label{sec:conclusion}
% \vspace{-0.1cm}
We present a parallel, GPU-friendly framework for self-supervised neural combinatorial optimization under constraints. Independent oracle queries generate feasible candidates in parallel, while a regularized geometric function supplies training gradients without requiring differentiation through the discrete oracle. Compared with earlier neural approaches, this design reduces sequential computation and produces feasible solutions directly at inference time, without rounding. Across MaxCut, Quadratic Assignment, Maximum Coverage, and Densest \(k\)-Subgraph, the framework achieves competitive solution quality with low inference latency. These results suggest a scalable approach to label-free learning for constrained combinatorial optimization. The framework relies on a well-characterized feasible polytope and an efficient exact or approximate linear minimization oracle (LMO), which may limit its applicability to less structured constraint families. Moreover, feasibility alone does not imply an approximation guarantee for the original objective. Extending the framework to broader classes of constraints and establishing corresponding approximation guarantees are important directions for future work.

\newpage

\subsection*{AI use statement}
We used generative AI tools to polish the writing and assist with implementing our method and experiments. We did \textbf{not} use AI to develop research ideas, formulate mathematical claims, write proofs, design experiments, or interpret results. The authors reviewed all AI-edited text, and one author reviewed and tested all AI-generated code for correctness. We take responsibility for the final content of this work,
including text, claims or artifacts produced with the aid of generative AI.

\subsection*{Reproducibility statement}
The appendix provides experimental details, including training data, hyperparameters, and baseline configurations. Code and data are available in  \href{https://github.com/frankx2023/Parallel-NCO}{Link}.

\bibliography{references}
\bibliographystyle{iclr2027_conference}

\appendix
\newpage

\section{Related work}
\label{sec:related}

\textbf{Self-supervised neural CO.}
Neural CO methods learn from distributions of problem instances using supervised learning, reinforcement learning, or objectives derived directly from the combinatorial problem
\citep{vinyals2015pointer,khalil2017learning,kool2019attention,amizadeh2018learning}.
Self-supervised methods use the discrete objective as a training signal without requiring optimal solutions as labels. They commonly optimize a continuous surrogate, often expressed as the expected objective value under a distribution parameterized by the network
\citep{karalias2020erdos,toenshoff2021graph,wang2022unsupervised,bu2024tackling}.
Besides the choice of neural architecture and input features
\citep{min2022can,sato2019approximation,lim2023sign,keriven2023functions},
the design of this surrogate and the recovery of a feasible solution are central parts of the solver.

Hard constraints complicate both steps. Penalty terms can discourage violations, but their weights may require tuning and a feasible output may still need to be recovered. Other methods use differentiable projections or specialized layers for cardinality, assignment, or more general linear constraints
\citep{donti2021dc3,wang2023linsatnet,zeng2024glinsat,sinkhorn1964relationship,wang2022towards}.
Annealing and physics-inspired formulations offer further ways to shape the training objective
\citep{sun2022annealed,schuetz2022combinatorial,mundinger2025neural}.
Geometric approaches instead define distributions supported on feasible solutions of a combinatorial polytope. The expected discrete objective then supplies a self-supervised loss, and the same feasible support can be used at inference
\citep{Karalias2022NeuralSF,karalias2025geometric,nerem2025differentiable,Rafiey2026-FWNCO}.
Our method follows this feasible-support perspective while constructing the support from independent oracle queries instead of building it through sequential decomposition.

\textbf{Parallel approaches to CO.}
Parallelization, especially on GPUs, has long been an important route to faster CO algorithms. GPUs can evaluate many states, candidate solutions, or problem instances concurrently, and their throughput is especially useful when computations can be expressed as batched vector and matrix operations. Examples include GPU-based Max Cut search, massively parallel optimization of binary programs, parallel Frank--Wolfe variants, and GPU kernels for batched dynamic programming
\citep{cook2019gpu,abbas2022fastdog,wang2016parallel,zhao2026sequential}.
Recent neural solvers also exploit GPU parallelism during prediction. Non-autoregressive graph models predict edge preferences for many decisions at once, while diffusion-based solvers update many variables simultaneously within each denoising step
\citep{joshi2019efficient,sun2023difusco}.
These approaches make the prediction stage more parallel, motivating closer attention to the subsequent steps that construct a feasible discrete solution.

Many neural CO methods use parallel neural computation to produce continuous predictions, such as heatmaps or points in a relaxation, but obtaining a high-quality feasible solution can require substantial sequential computation after the network output is available. Heatmap-based methods predict scores for many edges or variables simultaneously, then assemble solutions through beam search, conflict-aware greedy decoding, or heatmap-guided Monte Carlo tree search (MCTS)
\citep{joshi2019efficient,fu2021generalize,qiu2022dimes,sun2023difusco}.
Diffusion solvers also perform successive denoising steps before decoding
\citep{qiu2022dimes,sun2023difusco}.
Other learned predictors supply scores or heuristics to a subsequent combinatorial algorithm: NeuroLKH constructs candidate edge sets for LKH search, DeepACO guides ant-colony search
\citep{xin2021neurolkh,ye2023deepaco}.
Autoregressive neural solvers take another route, repeatedly selecting the next decision conditioned on the partial solution
\citep{kool2019attention,kwon2020pomo}.
Sequential dependence also appears in geometric methods: GeoNCO constructs vertices step by step within each decomposition, and FWNCO's next oracle query depends on the preceding decomposition step
\citep{karalias2025geometric,Rafiey2026-FWNCO}.
COExpander predicts heatmaps and determines variables over multiple adaptive expansion rounds, rather than completing solution construction in a single prediction step \citep{ma2025coexpander}.
Thus, although neural predictions can be computed in parallel, converting them into feasible solutions often requires sequential decoding, search, or decomposition. These steps can limit end-to-end speed, while solution quality may depend heavily on the choice of decoding or search procedure. Our method addresses the candidate-generation stage directly: independently generated oracle queries produce feasible solutions in parallel, which support the training extension and can be evaluated directly at inference.

\section{Proof of Proposition~\ref{prop:aediff}}
\label{app:proof-aediff}

\begin{proof}
We first fix the generated vertex matrix
$V=[\bv_1,\ldots,\bv_m]$ and consider the dependence on $\bx$.
The Hessian of the objective in \eqref{eq:alpha_projection}
with respect to $\balpha$ is
$V^\top V+\gamma I\succeq \gamma I\succ0$.
Hence the objective is strongly convex in $\balpha$, while the feasible
set $\{\balpha\geq0:\mathbf{1}^\top\balpha=1\}$ is convex and compact.
Therefore, for every $\bx$, \eqref{eq:alpha_projection} has a unique
minimizer $\balpha(\bx)$.

We next show that the solution map is Lipschitz for fixed $V$.
Let $\balpha=\balpha(\bx)$ and
$\widetilde{\balpha}=\balpha(\widetilde{\bx})$.
First-order optimality of the two convex problems over the same feasible
set gives
\begin{align*}
    \big\langle
    (V^\top V+\gamma I)\balpha-V^\top\bx,
    \widetilde{\balpha}-\balpha
    \big\rangle
    &\geq 0,\\
    \big\langle
    (V^\top V+\gamma I)\widetilde{\balpha}
    -V^\top\widetilde{\bx},
    \balpha-\widetilde{\balpha}
    \big\rangle
    &\geq 0.
\end{align*}
Adding the two inequalities and using
$V^\top V+\gamma I\succeq\gamma I$ yields
\begin{align*}
    \gamma
    \|\balpha-\widetilde{\balpha}\|_2^2
    &\leq
    \big\langle
    V^\top(\bx-\widetilde{\bx}),
    \balpha-\widetilde{\balpha}
    \big\rangle \\
    &\leq
    \|V\|_2
    \|\bx-\widetilde{\bx}\|_2
    \|\balpha-\widetilde{\balpha}\|_2.
\end{align*}
Thus,
\[
    \|\balpha(\bx)-\balpha(\widetilde{\bx})\|_2
    \leq
    \frac{\|V\|_2}{\gamma}
    \|\bx-\widetilde{\bx}\|_2.
\]
In particular, $\balpha(\bx)$ is continuous.

We now characterize the solution on a fixed positive support
$S=\{i:\balpha_i(\bx)>0\}$.
As derived in the main text, the positive coefficients and the multiplier
$\lambda$ satisfy \eqref{eq:kkt-positive}. The matrix $K_S$ is
nonsingular. Indeed, suppose
\[
    K_S
    \begin{bmatrix}
        \bu\\
        t
    \end{bmatrix}
    =0.
\]
Then $\mathbf{1}^\top\bu=0$ and
$(V_S^\top V_S+\gamma I)\bu+t\mathbf{1}=0$.
Multiplying the latter equality by $\bu^\top$ gives
\[
    \bu^\top(V_S^\top V_S+\gamma I)\bu=0.
\]
Since $V_S^\top V_S+\gamma I\succ0$, this implies $\bu=0$, and
consequently $t=0$. Hence $K_S$ is invertible and
\[
    \begin{bmatrix}
        \balpha_S(\bx)\\
        \lambda(\bx)
    \end{bmatrix}
    =
    K_S^{-1}
    \begin{bmatrix}
        V_S^\top\bx\\
        1
    \end{bmatrix}.
\]
Therefore, $\balpha_S(\bx)$ is affine in $\bx$ whenever the support $S$
is fixed.

For $i\notin S$, stationarity gives
\[
    \boldsymbol\mu_i
    =
    \bv_i^\top V_S\balpha_S(\bx)
    -\bv_i^\top\bx
    +\lambda(\bx),
\]
which is also affine in $\bx$. Thus, for each support $S$, the KKT
conditions $\balpha_S\geq0$ and
$\boldsymbol\mu_i\geq0$ for $i\notin S$ are affine inequalities in
$\bx$ and define a polyhedral region. Since there are finitely many
supports $S\subseteq[m]$, $\balpha(\bx)$ is piecewise affine.
Consequently, it is differentiable except on the boundaries of finitely
many polyhedral regions, which form a set of Lebesgue measure zero.

The same conclusion holds for
$F_V(\bx)=\sum_{i=1}^m\balpha_i(\bx)f(\bv_i)$, since the values
$f(\bv_i)$ are fixed when $V$ is fixed. On any region where the support
is fixed, differentiating \eqref{eq:kkt-positive} gives
\eqref{eq:implicit_jacobian}. Combining this with
\eqref{eq:extension_gradient} and the adjoint system
\eqref{eq:adjoint_kkt} gives
\[
    \nabla_{\bx}F_V(\bx)=V_S\bz.
\]
Hence the backward signal returned by
\Cref{alg:parallel-extension} is the exact gradient on each
fixed-support region, and therefore for a.e.\ $\bx$.

It remains to consider the dependence on the network parameters
$\theta$. By assumption, the deterministic LMO makes $V(\theta)$
well-defined, and $V(\theta)$ is locally constant for a.e.\ $\theta$.
Fix such a point and a neighborhood on which $V(\theta)=V$ is constant.
Since each $\bq_i(\theta)$ is locally Lipschitz,
$\bx_\theta=m^{-1}\sum_i\bq_i(\theta)$ is locally Lipschitz as well.
The Lipschitz bound above then implies that
$\balpha(\bx_\theta)$ is locally Lipschitz in $\theta$, and therefore so
is $F_V(\bx_\theta)$. By Rademacher's theorem,
$F_{V(\theta)}(\bx_\theta)$ is differentiable for a.e.\ $\theta$.

Finally, we verify that the backward computation in
\Cref{alg:parallel-extension} gives this derivative.
Consider a point $\theta$ at which $V$ is locally constant and
$\bx_\theta$, $\balpha(\bx_\theta)$, and the KKT multiplier $\lambda$
are differentiable; these conditions fail only on a measure-zero set.
Let
$S=\{i:\balpha_i(\bx_\theta)>0\}$.
For every $i\notin S$, the scalar function
$\balpha_i(\bx_\theta)$ is nonnegative and equals zero at the point
under consideration. Hence, at every point where it is differentiable,
$d\balpha_i=0$. For every $i\in S$, continuity implies
$\balpha_i>0$ in a neighborhood, so complementary slackness gives
$\boldsymbol\mu_i=0$ there. Differentiating the corresponding
stationarity equations together with
$\mathbf{1}^\top\balpha=1$ therefore gives
\[
    K_S
    \begin{bmatrix}
        \partial\balpha_S\\
        \partial\lambda
    \end{bmatrix}
    =
    \begin{bmatrix}
        V_S^\top \partial\bx_\theta\\
        0
    \end{bmatrix}.
\]
Here $\lambda$ is locally Lipschitz as well: whenever
$\balpha_i>0$, stationarity gives
$\lambda=\bv_i^\top\bx_\theta-
[(V^\top V+\gamma I)\balpha]_i$.

Since $V$ is locally constant, the values $f(\bv_i)$ are also constant,
and therefore
$\partial F=f_S^\top \partial\balpha_S$.
Let $(\bz,\eta)$ solve \eqref{eq:adjoint_kkt}. Since $K_S$ is symmetric,
\begin{align*}
    dF
    &=
    \begin{bmatrix}
        f_S^\top & 0
    \end{bmatrix}
    \begin{bmatrix}
        \partial\balpha_S\\
        \partial\lambda
    \end{bmatrix} \\
    &=
    \begin{bmatrix}
        \bz^\top & \eta
    \end{bmatrix}
    K_S
    \begin{bmatrix}
        \partial\balpha_S\\
        \partial\lambda
    \end{bmatrix} \\
    &=
    \bz^\top V_S^\top \partial\bx_\theta
    =
    (V_S\bz)^\top \partial\bx_\theta.
\end{align*}
Thus $V_S\bz$, exactly the signal returned by
\Cref{alg:parallel-extension}, gives the correct differential through
$\bx_\theta$ for a.e.\ $\theta$. Backpropagating this signal through
$\bx_\theta=m^{-1}\sum_i\bq_i(\theta)$ therefore computes the gradient
of $F_{V(\theta)}(\bx_\theta)$ with respect to $\theta$ a.e.
\end{proof}

\section{Hardware and VRAM usage}
\label{sec:hardware}

All our experiments except for the QAP baselines were conducted on a workstation equipped with an AMD Ryzen 9 9950X processor (16 cores, 32 threads) and an NVIDIA GeForce RTX 4090 GPU with 24\,GB of memory. The implementation uses Python 3.12.14 and PyTorch 2.12.1 with CUDA 13.2, running on Debian 13 under WSL2.  For QAP baselines, we use published results from previous studies. The reasons for this treatment are given in Appendix~\ref{app:qap-published-results}.
Table~\ref{tab:gpu-memory} reports peak GPU memory allocated by PyTorch during training and inference.
\begin{table}[H]
\small
\centering
\caption{Peak allocated GPU memory (GiB) of our method.}
\label{tab:gpu-memory}
\label{tab:training-memory}
\label{tab:inference-memory}
\label{tab:maxcut-training-vram}
\label{tab:dks-training-vram}
\label{tab:dks-inference-vram-largest}
\setlength{\tabcolsep}{6pt}
\resizebox{\linewidth}{!}{%
\begin{tabular}{llrr}
\toprule
Problem & Dataset / setting & Training (GiB) & Inference (GiB) \\
\midrule
MC & Random500 ($m=50$) & 0.081 & 0.014 \\
MC & Random1000 ($m=50$) & 0.160 & 0.024 \\
MC & Random5000 ($m=200$) & 0.581 & 0.120 \\
MC & Random10000 ($m=200$) & 1.141 & 0.228 \\
MC & Random50000 ($m=200$) & 5.596 & 1.094 \\
MC & Rail & -- & 0.225 \\
MC & Twitch & -- & 0.163 \\
\midrule
QAP & QAP32 ($m=5000$) & 2.951 & 0.111 \\
QAP & QAPLIB ($n=64$) & -- & 0.395 \\
\midrule
MaxCut & Mixed sizes ($n=250/500/1000$, $m=16$) & 1.059 & -- \\
\midrule
DKS & Synthetic ($n=500$, $m=32$) & 0.140 & -- \\
DKS & Synthetic ($n=500$, $m=200$) & 0.299 & -- \\
DKS & Orkut ($n=3{,}072{,}441$, $|E|=117{,}185{,}083$) & -- & 15.342 \\
DKS & LiveJournal ($n=4{,}847{,}571$, $|E|=42{,}851{,}237$) & -- & 22.082 \\
\bottomrule
\end{tabular}}
\par\smallskip
\begin{minipage}{\linewidth}\footnotesize
For MC and QAP, training batch sizes are 4 and 16, respectively; inference batch size is 1.
\end{minipage}
\end{table}

\section{Details and Further Experiments on MaxCut}
\label{sec:appendix-maxcut}
\subsection{Experiment Setup}
\label{app:maxcut-setup}

\paragraph{Training and test graphs.}
Following ECO-DQN~\citep{barrett2020exploratory} and
ANYCSP~\citep{tonshoff2022one}, we train on generated random graphs and
evaluate transfer to established benchmarks. ECO-DQN uses
Erd\H{o}s--R\'enyi and Barab\'asi--Albert graphs, and ANYCSP uses
Erd\H{o}s--R\'enyi graphs for MaxCut. We extend this approach by mixing
four unweighted graph families:
Erd\H{o}s--R\'enyi, random regular, Barab\'asi--Albert, and periodic
two-dimensional grids with random shortcuts. Training interleaves batches of
$n\in\{250,500,1000\}$, with one graph size per batch.
For testing, we use Gset, an established MaxCut benchmark collection
maintained by Yinyu Ye\footnote{\url{https://web.stanford.edu/~yyye/yyye/Gset/}}.
Following the instance selection and size grouping of
ANYCSP~\citep{tonshoff2022one}, we evaluate 35 unweighted graphs:
nine graphs each with 800, 1000, and 2000
vertices, and eight graphs with 3000--10000 vertices. We report mean cut
values and per-graph runtimes; the overall mean weights all 35 graphs equally.

\paragraph{Graph representation and neural network.}
Each vertex receives 16 breadth-first-search (BFS) parity features and 16 random
normal features. A parity feature records whether the shortest-path distance
from a sampled reference vertex is even or odd, with zero for unreachable
vertices\citep{li2020distance}. We use a width-32 residual message-passing network with 512 layers and produces an embedding matrix.
The neural network combines this embedding with fixed Sobol vectors to generate $m$ query vectors:
\[
 q_j=H_\theta\bigl(c_\theta(d_j)\odot z_j\bigr),\qquad
 c_\theta(d)=0.75+0.5\,\operatorname{sigmoid}(d)
                  +0.2\tanh(W_qd),
\]
where $(d_j,z_j)$ is a pair of fixed input vectors.

\paragraph{Training and inference.}
We train for 30 epochs with 64 new graphs per epoch, batch size 4, and
$m_{\mathrm{train}}=16$. AdamW uses learning rate $10^{-3}$, weight decay
$10^{-5}$, and gradient clipping at norm 1.
At inference, we score all $m$ assignments and return the largest cut.
The main results use $m=1000$.

\subsection{Related Work and Baselines}
\label{app:maxcut-baselines}

The traditional baselines are Greedy and semidefinite programming (SDP).
The SDP relaxation follows Goemans and Williamson~\citep{goemans1995improved}.
RUN-CSP, ECO-DQN, and ANYCSP are neural-network methods.
RUN-CSP~\citep{toenshoff2021graph} uses recurrent message passing with
unsupervised training. ECO-DQN~\citep{barrett2020exploratory} uses a graph
neural network and deep Q-learning to select vertex flips.
ANYCSP~\citep{tonshoff2022one} uses a graph neural network trained with
policy gradients to update complete assignments.
Our comparison uses the implementations and bundled ER800 pretrained
checkpoints in MaxCut-Bench\footnote{\url{https://github.com/ankurnath/MaxCut-Bench}}.
Following the multi-run evaluation approach of ANYCSP~\citep{tonshoff2022one},
we use 20 trajectories per neural baseline and a 180-second limit for ANYCSP, set RUN-CSP to 1600 recurrent steps and ECO-DQN to $2n$ steps. SDP is evaluated on the 800- and 1000-vertex groups with a three-hour limit per graph.
Time columns report the recorded mean runtimes.

\subsection{Scaled Gaps and Raw Cut Values}
Table~\ref{tab:maxcut-raw-cuts} provides the raw cuts corresponding to the scaled gaps in Table~\ref{tab:completed-cut}. Both tables weight all graphs equally within each group and across all 35 instances.

\begin{table*}[h]
\centering
\caption{MaxCut results on 35 Gset graphs. Mean gap (\%) and time: lower is better. \textcolor{Red!50}{$\blacksquare$} and \textcolor{Blue!50}{$\blacksquare$} indicate the best and second-best results, respectively.}
\label{tab:completed-cut}
\setlength{\tabcolsep}{5pt}
\resizebox{\textwidth}{!}{%
\begin{tabular}{lrrrrrrrrrr}
\toprule
& \multicolumn{2}{c}{\textbf{Gset800 (9 graphs)}} & \multicolumn{2}{c}{\textbf{Gset1000 (9 graphs)}} & \multicolumn{2}{c}{\textbf{Gset2000 (9 graphs)}} & \multicolumn{2}{c}{\textbf{Gset3000+ (8 graphs)}} & \multicolumn{2}{c}{\textbf{Mean (35 graphs)}} \\
\cmidrule(lr){2-3}\cmidrule(lr){4-5}\cmidrule(lr){6-7}\cmidrule(lr){8-9}\cmidrule(lr){10-11}
\textbf{Method} & Gap (\%) $\downarrow$ & Time (s) $\downarrow$ & Gap (\%) $\downarrow$ & Time (s) $\downarrow$ & Gap (\%) $\downarrow$ & Time (s) $\downarrow$ & Gap (\%) $\downarrow$ & Time (s) $\downarrow$ & Gap (\%) $\downarrow$ & Time (s) $\downarrow$ \\
\midrule
Greedy & 2.93 & \rkone{0.0009} & 3.74 & \rkone{0.0010} & 3.75 & \rkone{0.0029} & 3.51 & \rkone{0.0190} & 3.48 & \rkone{0.0056} \\
Gurobi (1s) & 0.96 & 1.0378 & 1.20 & 1.0319 & 1.62 & 1.0368 & 1.35 & 0.8349 & 1.28 & 0.9896 \\
SDP & 3.22 & 502.1552 & 3.76 & 1929.5763 & -- & -- & -- & -- & -- & -- \\
RUN-CSP & 3.06 & 5.9356 & 3.22 & 4.4649 & 3.31 & 8.1989 & 2.47 & 10.6793 & 3.03 & 7.2237 \\
ECO-DQN & 9.86 & 16.0096 & 10.76 & 20.8582 & 11.69 & 48.7131 & 15.68 & 177.0220 & 11.89 & 62.4687 \\
ANYCSP (180s) & \rktwo{0.49} & 180.1859 & \rktwo{0.80} & 180.2433 & \rktwo{0.96} & 180.1814 & 1.61 & 138.1003 & 0.95 & 170.5800 \\
BLS (1s) & \rkone{0.12} & 1.0000 & \rkone{0.15} & 1.0000 & \rkone{0.24} & 1.0000 & \rkone{0.41} & 1.0000 & \rkone{0.22} & 1.0000 \\
\midrule
Ours & 0.88 & \rktwo{0.2509} & 0.99 & \rktwo{0.0707} & 1.00 & \rktwo{0.3002} & \rktwo{0.58} & \rktwo{0.2063} & \rktwo{0.87} & \rktwo{0.2070} \\
\bottomrule
\end{tabular}}
\par\smallskip
\begin{minipage}{\textwidth}\footnotesize
We are using the BLS\citep{benlic2013breakout} reference cuts and averaging equally over graphs. Raw cut values are provided in Table~\ref{tab:maxcut-raw-cuts}.
\end{minipage}
\end{table*}

\begin{table}[h]
\centering
\caption{Raw MaxCut values and inference time on 35 Gset graphs. Mean cut: higher is better; time: lower is better.}
\label{tab:maxcut-raw-cuts}
\setlength{\tabcolsep}{5pt}
\resizebox{\linewidth}{!}{%
\begin{tabular}{lrrrrrrrrrr}
\toprule
& \multicolumn{2}{c}{Gset800 (9 graphs)} & \multicolumn{2}{c}{Gset1000 (9 graphs)} & \multicolumn{2}{c}{Gset2000 (9 graphs)} & \multicolumn{2}{c}{Gset3000+ (8 graphs)} & \multicolumn{2}{c}{Mean (35 graphs)} \\
\cmidrule(lr){2-3}\cmidrule(lr){4-5}\cmidrule(lr){6-7}\cmidrule(lr){8-9}\cmidrule(lr){10-11}
Method & Cut $\uparrow$ & Time (s) & Cut $\uparrow$ & Time (s) & Cut $\uparrow$ & Time (s) & Cut $\uparrow$ & Time (s) & Cut $\uparrow$ & Time (s) \\
\midrule
Greedy & 7630.89 & 0.0009 & 5212.56 & 0.0010 & 10433.00 & 0.0029 & 11748.25 & 0.0190 & 8670.69 & 0.0056 \\
Gurobi (1s) & 7774.00 & 1.0378 & 5349.00 & 1.0319 & 10664.89 & 1.0368 & 12053.62 & 0.8349 & 8872.00 & 0.9896 \\
SDP & 7594.00 & 502.1552 & 5205.33 & 1929.5763 & -- & -- & -- & -- & -- & -- \\
RUN-CSP & 7625.11 & 5.9356 & 5250.56 & 4.4649 & 10500.33 & 8.1989 & 11855.75 & 10.6793 & 8720.86 & 7.2237 \\
ECO-DQN & 7432.44 & 16.0096 & 4948.89 & 20.8582 & 9820.56 & 48.7131 & 10194.62 & 177.0220 & 8039.26 & 62.4687 \\
ANYCSP (180s) & 7800.00 & 180.1859 & 5368.56 & 180.2433 & 10730.33 & 180.1814 & 12048.62 & 138.1003 & 8899.40 & 170.5800 \\
BLS (1s) & 7810.70 & 1.0000 & 5401.22 & 1.0000 & 10804.44 & 1.0000 & 12212.79 & 1.0000 & 8967.13 & 1.0000 \\
\midrule
Ours & 7751.07 & 0.2509 & 5353.74 & 0.0707 & 10719.04 & 0.3002 & 12174.29 & 0.2063 & 8908.83 & 0.2070 \\
\bottomrule
\end{tabular}}
\end{table}

\subsection{Effect of the Parallel Query Budget}
\label{app:maxcut-query-budget}

The \emph{parallel query budget} $m$ is the number of query directions and
corresponding feasible candidates generated by the neural network.
We keep the trained weights fixed and evaluate
$m\in\{200,500,1000,2000,5000\}$. For budgets above 1000, we extend the same
fixed Sobol sequence, preserving the original vector prefix.
Each budget is timed with a separate full forward pass, including query
generation, sign decoding, scoring, and selection; graph loading and BFS
preprocessing are outside this inference measurement.

Table~\ref{tab:maxcut-m-ablation} shows modest gains as $m$ increases: the mean
cut rises from 8907.05 at $m=200$ to 8910.09 at $m=5000$.
The measured mean inference time changes only from 0.0545 to 0.0562 seconds,
consistent with sharing the graph encoder across all queries.
The main setting $m=1000$ achieves 8908.83, already close to the largest-budget
result. The same trained model supports all five inference budgets.

\begin{table}[H]
\centering
\caption{Effect of the parallel query budget $m$ on MaxCut. Mean cut and inference time over three repetitions.}
\label{tab:maxcut-m-ablation}
\resizebox{\linewidth}{!}{\begin{tabular}{lrrrrrrrrrrr}
\toprule
Method & $m$ & \multicolumn{2}{c}{Gset800} & \multicolumn{2}{c}{Gset1000} & \multicolumn{2}{c}{Gset2000} & \multicolumn{2}{c}{Gset3000+} & \multicolumn{2}{c}{All 35} \\
 & & Cut & Time (s) & Cut & Time (s) & Cut & Time (s) & Cut & Time (s) & Cut & Time (s) \\
\midrule
Ours & 200 & 7749.67 & 0.0544 & 5352.44 & 0.0548 & 10716.59 & 0.0549 & 12172.29 & 0.0537 & 8907.05 & 0.0545 \\
Ours & 500 & 7750.59 & 0.0542 & 5353.19 & 0.0547 & 10717.96 & 0.0549 & 12173.46 & 0.0535 & 8908.10 & 0.0543 \\
Ours & 1000 & 7751.07 & 0.0550 & 5353.74 & 0.0546 & 10719.04 & 0.0556 & 12174.29 & 0.0537 & 8908.83 & 0.0547 \\
Ours & 2000 & 7751.70 & 0.0549 & 5354.44 & 0.0551 & 10719.89 & 0.0555 & 12174.54 & 0.0543 & 8909.45 & 0.0550 \\
Ours & 5000 & 7752.15 & 0.0557 & 5354.93 & 0.0560 & 10721.11 & 0.0569 & 12174.92 & 0.0563 & 8910.09 & 0.0562 \\
\bottomrule
\end{tabular}}
\end{table}

\section{Details and Further Experiments on QAP}
\label{sec:appendix-qap}

\subsection{Experiment Setup}
\label{app:qap-setup}

We follow previous approaches prepare data on synthetic datasets and evaluate on the QAPLIB dataset.
Training uses the same synthetic QAP32 dataset, with 5120 training instances generated at edge probability $p=0.3$ and 256 held-out instances. The neural network processes the flow and distance matrices using the matrix features and GraphSAGE representation. It predicts Gaussian mean and variance matrices, from which queries are sampled.

We evaluate on the 14 QAPLIB classes with instance sizes
$12\leq n\leq64$. The gap is
$100(C/C_{\mathrm{ref}}-1)$, where $C$ is the predicted cost and
$C_{\mathrm{ref}}$ is the QAPLIB reference value.
We report per-class mean gaps, their average and mean inference time per instance in Table~\ref{tab:qap-mean-time}.

\begin{table*}[t]
\small
\centering
\setlength{\tabcolsep}{3pt}
\caption{We report the optimality gap (\%; lower is better) for each QAPLIB class,
the average gap across classes, and mean inference time.
\textcolor{Red!50}{$\blacksquare$} and
\textcolor{Blue!50}{$\blacksquare$} indicate the best and second-best
results, respectively.}
\label{tab:qap-mean-time}
\resizebox{\textwidth}{!}{%
\begin{tabular}{l|cccccccccccccc|c|c}
\toprule
& \multicolumn{15}{c|}{\textbf{Gap (\%)} $\downarrow$} & \\
\cmidrule(lr){2-16}
\textbf{Method} & \textbf{bur} & \textbf{chr} & \textbf{esc} & \textbf{had} & \textbf{kra} & \textbf{lipa} & \textbf{nug} & \textbf{rou} & \textbf{scr} & \textbf{sko} & \textbf{ste} & \textbf{tai} & \textbf{tho} & \textbf{wil} & \textbf{Avg.} & \textbf{Time (s)} $\downarrow$ \\
\midrule
RRWM & 23.1 & 616.0 & 63.9 & 25.1 & 58.8 & 20.9 & 67.8 & 51.2 & 173.5 & 48.5 & 539.4 & 197.2 & 80.6 & 18.2 & 169.5 & \rktwo{0.150} \\
SK-JA & 4.7 & \rktwo{38.5} & 364.8 & 25.8 & 41.4 & \rkone{0.0} & 25.3 & 13.7 & 48.6 & 18.3 & 120.4 & 25.2 & 32.9 & 8.8 & 93.2 & 563.400 \\
NGM & 3.4 & 121.3 & 126.7 & 8.2 & 31.6 & 16.2 & 21.0 & 30.9 & 55.5 & 25.2 & 101.7 & 61.4 & 27.5 & 10.8 & 62.4 & 15.720 \\
RGM & 7.1 & 112.4 & 32.8 & 6.2 & \rktwo{15.0} & 13.3 & 9.7 & 13.4 & 45.5 & \rktwo{10.6} & 134.1 & 17.3 & 20.7 & 8.1 & 35.8 & 75.530 \\
SAWT & 2.8 & 110.7 & \rktwo{13.5} & 3.8 & 30.1 & \rktwo{0.4} & \rktwo{9.2} & 10.8 & 28.5 & 17.7 & 93.5 & \rktwo{16.5} & 24.8 & 8.1 & 26.8 & 12.110 \\
FWNCO & \rktwo{2.3} & 97.7 & 19.0 & \rktwo{2.8} & 37.0 & 13.5 & 11.5 & 10.9 & 23.6 & 15.0 & 83.5 & 20.7 & 23.0 & 9.2 & \rktwo{26.4} & 1.560 \\
RoTS (1s) & \rkone{0.0} & \rkone{0.2} & \rkone{0.0} & \rkone{0.0} & \rkone{0.0} & \rkone{0.0} & \rkone{0.0} & \rkone{0.0} & \rkone{0.0} & \rkone{0.1} & \rkone{0.0} & \rkone{0.2} & \rkone{0.0} & \rkone{0.0} & \rkone{0.0} & 1.000 \\
\midrule
Ours & 2.4 & 127.6 & 18.5 & 3.2 & 33.6 & 13.3 & 10.1 & \rktwo{9.5} & \rktwo{19.7} & 13.5 & \rktwo{66.4} & 23.6 & \rktwo{20.6} & \rktwo{7.6} & \rktwo{26.4} & \rkone{0.036} \\
\bottomrule
\end{tabular}%
}
\end{table*}

\subsection{Baselines}
\label{app:qap-baselines}

\paragraph{Classical methods.}
SM~\citep{leordeanu2005spectral} uses the principal eigenvector of the affinity
matrix to score candidate correspondences, followed by discrete matching.
RRWM~\citep{cho2010reweighted} performs random walks on an association graph
and reweights correspondence scores to enforce matching constraints.
SK-JA~\citep{kushinsky2019sinkhorn} solves the Johnson--Adams relaxation of
QAP using generalized Sinkhorn iterations and Bregman projections. \rev{Robust Tabu Search (RoTS)~\citep{Taillard1991} searches over assignment
permutations using pairwise swaps, with tabu memory to discourage cycling
and aspiration criteria to allow promising tabu moves.
We use Taillard's publicly released C implementation as a non-neural
metaheuristic baseline.}

\paragraph{Neural methods.}
NGM~\citep{wang2021neural} learns node embeddings on the association graph
and predicts a matching with Sinkhorn normalization.
RGM~\citep{liu2023revocable} uses deep reinforcement learning to construct
node correspondences sequentially, with revocable actions that allow
earlier matching decisions to be revised.
SAWT~\citep{tan2024learning} uses a solution-aware Transformer to encode
the current assignment and learns actions that improve its QAP cost.
FWNCO~\citep{Rafiey2026-FWNCO} predicts a continuous assignment matrix and applies
sequential Frank--Wolfe decomposition with a greedy assignment oracle,
returning the generated permutation with the lowest objective value.

\subsection{Source of QAPLIB Baseline Results}
\label{app:qap-published-results}

We use the published QAPLIB baseline results from previous works. Reusing reported results is established practice in QAP evaluation: SAWT~\citep{tan2024learning} follows RGM's~\citep{liu2023revocable}
QAPLIB comparison protocol and reuses previously published baseline results.

As discussed in FWNCO~\citep[Appendix~C.3]{Rafiey2026-FWNCO}, the public SAWT and RGM repositories do not provide the complete settings and evaluation pipeline needed to reproduce their reported QAPLIB experiments. That appendix also identifies inconsistencies in earlier QAPLIB summaries, including the WIL class, whose reported values cannot be reconciled with the stated instance selection. Reconstructing these experiments would therefore require unverified choices about the original evaluation. We retain the published baseline values to preserve the original comparisons, and evaluate Our method locally on QAPLIB.

\paragraph{Hardware and computational efficiency.}
To assess how much of our QAP efficiency advantage comes from the algorithm itself, we compare the hardware configurations used by previous methods.The reported platforms use an RTX 3080 Ti for SAWT~\citep{tan2024learning}, an RTX 8000 for NGM~\citep{wang2021neural} and FWNCO~\citep{Rafiey2026-FWNCO}, and an RTX 2080 Ti for RGM~\citep{liu2023revocable}; our experiments use an RTX 4090. These GPUs differ substantially in performance: nominal peak FP32 throughput is approximately 34, 16.3, 14.2, and 82.6 TFLOPS for the RTX 3080 Ti, RTX 8000, RTX 2080 Ti, and RTX 4090, respectively.
Meanwhile, the reported configurations for methods other than SAWT also use powerful workstation- or server-class CPUs and larger system memory capacities, which can affect runtime and the scale of instances they can process.

Table~\ref{tab:qap-mean-time} reports 0.036 seconds per instance for Our method, compared with 1.560, 12.110, 15.720, and 75.530 seconds for FWNCO, SAWT, NGM, and RGM. As an illustrative sensitivity analysis, even granting each baseline a hypothetical $10\times$ reduction in its entire reported runtime leaves time ratios of $4.3\times$, $33.6\times$, $43.7\times$, and $209.8\times$, respectively, relative to Ours. This supports a substantial
algorithmic efficiency advantage.

Our memory measurements in Table~\ref{tab:gpu-memory} also show a modest
footprint: 2.951 GiB of peak allocated GPU memory for QAP32 training
and 0.395 GiB for QAPLIB inference at $n=64$, using $m=5000$.

\subsection{Effect of the Parallel Query Budget}
\label{app:qap-query-budget}

We compare the \emph{parallel query budget}
$m\in\{100,200,500,1000,3000,5000,10000,20000\}$.
For QAP, this budget is implemented with $m$ learned Gaussian mean/variance
matrix pairs.
Evaluation draws four samples from each distribution, yielding $4m$
assignment directions per instance.
Table~\ref{tab:qap-m-ablation} shows that increasing $m$ reduces the average
gap from 35.11\% to 25.10\%, while inference time increases from 0.01248
to 0.11105 seconds. We use $m=5000$ in the main comparison, achieving a
26.39\% average gap in this experiment. Increasing the budget to $m=20000$
reduces the gap by a further 1.29 percentage points at approximately
$3.2\times$ the inference time.

\begin{table}[H]
\small
\centering
\caption{Effect of the parallel query budget $m$ on QAPLIB, with four samples per distribution. Gap and time: lower is better.}
\label{tab:qap-m-ablation}
\begin{tabular}{rrr}
\toprule
$m$ & Avg. gap (\%) & Time (s) \\
\midrule
100 & 35.11 & 0.01248 \\
200 & 32.17 & 0.01251 \\
500 & 29.82 & 0.01295 \\
1000 & 28.99 & 0.01507 \\
3000 & 26.97 & 0.02465 \\
5000 & 26.39 & 0.03425 \\
10000 & 25.75 & 0.05935 \\
20000 & 25.10 & 0.11105 \\
\bottomrule
\end{tabular}
\end{table}

% Required preamble packages: \usepackage{booktabs,graphicx,float}
\section{Details and Further Experiments on Maximum Coverage}
\label{sec:appendix-mc}

\subsection{Experiment Setup}
\label{app:mc-setup}

We follow GeoNCO~\citep{karalias2025geometric}for data preparation and evaluation on
Random500, Random1000, Rail, and Twitch, using the same datasets and
cardinality budgets. We report normalized coverage and mean inference
time per instance in Table~\ref{tab:mc-raw-coverage}.

\paragraph{Neural network training and inference.}
We represent each instance as a bipartite graph connecting sets to the
elements they contain, with node features based on degrees and element
weights. The model uses GraphSAGE network with four
bidirectional message-passing stages, hidden dimension 64, LeakyReLU activations, and dropout 0.1, it produces $m$ diagonal Gaussian distributions that is later used to sample query vectors.

We train separate models for each Random dataset and cardinality budget. For Random500 and Random1000, training uses AdamW with batch size 4, peak learning rate $10^{-3}$, weight decay $10^{-4}$, and a 20-epoch warmup followed by cosine decay to $5\times10^{-5}$.
During inference, we draw one query vector from each distribution produced by the neural network, run the LMO, and evaluate
the weighted coverage of all $m$ candidates in parallel. We return the candidate with the highest coverage.

\paragraph{No-TTO setting.}
We follow the no Test-time optimization (TTO) setting of FWNCO~\citep[Appendix~E]{Rafiey2026-FWNCO}, using
the short or no-TTO settings of the corresponding MC neural baselines.
Test-time optimization can improve solution quality through per-instance
fine-tuning or additional search, but it also adds computation and makes
the contribution of the learned model harder to isolate. \cite{xia2024position} discuss this issue for heatmap-guided search
in large-scale TSP, questioning how much the learned predictions contribute
to the final solution. Our evaluation focuses on solution generation with
fixed network parameters and the method's prescribed decoder.

\subsection{Baselines}
\label{app:mc-baselines}

We compare against exact solvers, classical heuristics, and neural methods.

\paragraph{Exact solvers.}
Gurobi solves a mixed-integer programming formulation with a 120-second
time limit per instance. We report the best feasible solution found within
the limit.

\paragraph{Classical heuristics.}
Random samples subsets of size $k$ uniformly and returns the best solution
found within 240 seconds. Greedy~\citep{NemhauserWF78} repeatedly adds the
set with the largest marginal coverage gain. For nonnegative element
weights, it achieves a $(1-1/\mathrm{e})$ approximation under the cardinality
constraint.
Following the discussion in GeoNCO~\citep{karalias2025geometric} , we also classify UCOM2~\citep{bu2024tackling} as a traditional baseline because its solution quality is primarily driven by incremental greedy derandomization, and we only report its short setting according to the no-TTO setting.

\paragraph{Neural methods.}
CardNN~\citep{wang2022towards} uses differentiable cardinality-constrained
learning; we report CardNN-noTTO-S using the evaluation setting of
Bu et al.~\citep{bu2024tackling}.
EGN-naive~\citep{karalias2020erdos} uses a probabilistic objective with
iterative rounding.
GeoNCO~\citep{karalias2025geometric} projects neural predictions onto the
feasible polytope and uses Carath\'eodory decomposition to produce feasible
solutions.
FWNCO~\citep{Rafiey2026-FWNCO} uses sequential Frank--Wolfe decomposition to generate
feasible subsets from a neural prediction and returns the subset with the
highest coverage.

\subsection{Very Large Graphs}
\label{app:mc-large}

We further evaluate on Random10000 and Random50000, containing 10000/50000
sets and 20000/100000 elements, respectively. Models are trained for
50 epochs with $m=200$, and each setting is evaluated on 100 test instances.
Table~\ref{tab:mc-random10000-random50000} compares Ours with the Greedy
implementation from UCom2. Across these settings, Ours achieves
94.8--96.6\% of Greedy's coverage using only approximately
0.01--0.33\% of its reported runtime.
The independent directions and top-$k$ oracle calls support
batched GPU execution even at these larger sizes.

\begin{table}[ht]
\small
\centering
\setlength{\tabcolsep}{4pt}
\caption{MC results on Random10000 and Random50000 ($m=200$ for Ours). Coverage: higher is better; time: lower is better.}
\label{tab:mc-random10000-random50000}
\resizebox{\linewidth}{!}{%
\begin{tabular}{l rrrr rrrr}
\toprule
& \multicolumn{4}{c}{\textbf{Random10000}} & \multicolumn{4}{c}{\textbf{Random50000}} \\
\cmidrule(lr){2-5}\cmidrule(lr){6-9}
& \multicolumn{2}{c}{$k=200$} & \multicolumn{2}{c}{$k=1000$}
& \multicolumn{2}{c}{$k=1000$} & \multicolumn{2}{c}{$k=5000$} \\
\cmidrule(lr){2-3}\cmidrule(lr){4-5}\cmidrule(lr){6-7}\cmidrule(lr){8-9}
Method & Coverage & Time (s) & Coverage & Time (s) & Coverage & Time (s) & Coverage & Time (s) \\
\midrule
Greedy & 311016.46 & 1.94177 & 886105.48 & 8.10018 & 1556362.95 & 51.06446 & 4432881.43 & 252.55409 \\
\midrule
Ours & 300336.75 & 0.00633 & 845122.08 & 0.00619 & 1479678.05 & 0.02492 & 4203456.14 & 0.02515 \\
\bottomrule
\end{tabular}%
}
\end{table}

\subsection{Effect of the Parallel Query Budget}
\label{app:mc-query-budget}

We compare the \emph{parallel query budget}
$m\in\{10,30,50,100,200\}$ on Random1000 with $k=20$.
Each budget uses a separately trained model with 30 training epochs and
the same training and test datasets. Here, the parallel query budget $m$
is the number of query vectors $q_1,\ldots,q_m$ generated for each instance.
Table~\ref{tab:mc-m-ablation} shows that coverage increases from 29506.38
at $m=10$ to 29878.97 at $m=50$. Increasing $m$ from 50 to 200 improves
coverage by only 0.43\%, with similar measured inference times due to parallel processing on the GPU.
We use $m=50$ for the main comparison because it achieves coverage close
to the larger budgets while generating fewer candidates. The very-large-graph
experiments use $m=200$ to provide a larger candidate set.

\begin{table}[H]
\small
\centering
\caption{Effect of the parallel query budget $m$ on MC on Random1000 ($k=20$).}
\label{tab:mc-m-ablation}
\begin{tabular}{rrr}
\toprule
$m$ & Coverage $\uparrow$ & Time (s) $\downarrow$ \\
\midrule
10  & 29506.38 & 0.00350 \\
30  & 29683.72 & 0.00359 \\
50  & 29878.97 & 0.00354 \\
100 & 29939.88 & 0.00363 \\
200 & 30008.05 & 0.00364 \\
\bottomrule
\end{tabular}
\end{table}

\subsection{Raw Coverage Values}

\begin{table*}[ht]
\centering
\caption{MC results on Random500, Random1000, Rail, and Twitch. Normalized coverage: higher is better; time: lower is better. \textcolor{Red!50}{$\blacksquare$} and \textcolor{Blue!50}{$\blacksquare$} indicate the best and second-best results among neural methods in each column, respectively.}
\label{tab:max-coverage-results}
\setlength{\tabcolsep}{3pt}
\resizebox{\textwidth}{!}{%
\begin{tabular}{l cccc cccc cccc cccc}
\toprule
& \multicolumn{4}{c}{\textbf{Random500}} & \multicolumn{4}{c}{\textbf{Random1000}} & \multicolumn{4}{c}{\textbf{Rail}}  & \multicolumn{4}{c}{\textbf{Twitch}} \\
\cmidrule(r){2-5} \cmidrule(lr){6-9} \cmidrule(l){10-13} \cmidrule(l){14-17}
& \multicolumn{2}{c}{$k=10$} & \multicolumn{2}{c}{$k=50$} & \multicolumn{2}{c}{$k=20$} & \multicolumn{2}{c}{$k=100$} & \multicolumn{2}{c}{$k=20$} & \multicolumn{2}{c}{$k=50$}  & \multicolumn{2}{c}{$k=20$} & \multicolumn{2}{c}{$k=50$} \\
\cmidrule(r){2-3} \cmidrule(r){4-5} \cmidrule(lr){6-7} \cmidrule(lr){8-9} \cmidrule(lr){10-11} \cmidrule(l){12-13} \cmidrule(lr){14-15} \cmidrule(l){16-17}
\textbf{Method} & Norm. cov. $\uparrow$ & Time (s) $\downarrow$ & Norm. cov. $\uparrow$ & Time (s) $\downarrow$ & Norm. cov. $\uparrow$ & Time (s) $\downarrow$ & Norm. cov. $\uparrow$ & Time (s) $\downarrow$ & Norm. cov. $\uparrow$ & Time (s) $\downarrow$ & Norm. cov. $\uparrow$ & Time (s) $\downarrow$ & Norm. cov. $\uparrow$ & Time (s) $\downarrow$ & Norm. cov. $\uparrow$ & Time (s) $\downarrow$ \\
\midrule
Random (240s) & 0.86 & 240.00019 & 0.82 & 240.00020 & 0.78 & 240.00009 & 0.79 & 240.00012 & 0.94 & 240.00103 & 0.97 & 240.00089 & 0.54 & 240.00028 & 0.54 & 240.00022 \\
Gurobi (120s) & 1.00 & 0.93991 & 1.00 & 120.07712 & 1.00 & 21.62572 & 1.00 & 120.05282 & 1.00 & 120.72583 & 1.00 & 120.62132 & 1.00 & 0.56769 & 1.00 & 0.53344 \\
Greedy & 0.99 & 0.00721 & 0.99 & 0.02072 & 0.99 & 0.02318 & 0.99 & 0.07883 & 1.00 & 0.07212 & 1.00 & 0.13551 & 1.00 & 0.06675 & 1.00 & 0.12559 \\
Ucom2-short & 0.98 & 0.45178 & 0.99 & 0.42560 & 0.96 & 0.84422 & 0.99 & 0.84090 & 0.98 & 1.42266 & 1.00 & 1.21777 & 1.00 & 8.24705 & 1.00 & 7.75699 \\
\midrule
GeoNCO & 0.96 & 0.96187 & 0.91 & 2.23917 & 0.95 & 1.28572 & 0.91 & 3.72534 & \rktwo{0.95} & 2.68297 & \rkone{0.97} & 4.04408 & \rktwo{0.32} & 2.44298 & \rktwo{0.37} & 3.21277 \\
CardNN-noTTO-S & 0.59 & 1.33321 & 0.73 & 1.32723 & 0.59 & 1.38980 & 0.73 & 1.35915 & 0.90 & \rktwo{0.94137} & 0.94 & \rktwo{0.99340} & 0.00 & \rktwo{0.66294} & 0.01 & \rktwo{0.62554} \\
EGN-naive & \rkone{0.97} & 53.041 & 0.92 & 120.136 & \rkone{0.96} & 120.010 & 0.90 & 120.356 & 0.93 & 120.676 & \rktwo{0.97} & 121.335 & 0.01 & 108.30300 & 0.01 & 105.46050 \\
FWNCO & \rktwo{0.96} & \rktwo{0.36310} & \rktwo{0.94} & \rktwo{0.35790} & 0.95 & \rktwo{0.39220} & \rktwo{0.93} & \rktwo{0.57820} & \rkone{0.95} & 1.00100 & 0.97 & 1.27860 & 0.32 & 0.68350 & 0.36 & 0.81810 \\
\midrule
Ours & 0.96 & \rkone{0.00383} & \rkone{0.94} & \rkone{0.00353} & \rktwo{0.95} & \rkone{0.00350} & \rkone{0.94} & \rkone{0.00352} & 0.94 & \rkone{0.00846} & 0.97 & \rkone{0.01191} & \rkone{0.39} & \rkone{0.00910} & \rkone{0.53} & \rkone{0.00842} \\
\bottomrule
\end{tabular}%
}
\par\smallskip
\begin{minipage}{\textwidth}\footnotesize
Normalized coverage is the mean coverage divided by the highest mean coverage in the same column. Raw coverage values are provided in Table~\ref{tab:mc-raw-coverage}.
\end{minipage}
\end{table*}

Table~\ref{tab:mc-raw-coverage} provides the unnormalized coverage values used in Table~\ref{tab:max-coverage-results}. Each main-table coverage column is divided by its largest entry.

\begin{table}[ht]
\centering
\caption{Raw MC coverage and inference time on Random500, Random1000, Rail, and Twitch. Coverage: higher is better; time: lower is better.}
\label{tab:mc-raw-coverage}
\setlength{\tabcolsep}{3pt}
\resizebox{\linewidth}{!}{%
\begin{tabular}{l cccc cccc cccc cccc}
\toprule
& \multicolumn{4}{c}{\textbf{Random500}} & \multicolumn{4}{c}{\textbf{Random1000}} & \multicolumn{4}{c}{\textbf{Rail}}  & \multicolumn{4}{c}{\textbf{Twitch}} \\
\cmidrule(r){2-5} \cmidrule(lr){6-9} \cmidrule(l){10-13} \cmidrule(l){14-17}
& \multicolumn{2}{c}{$k=10$} & \multicolumn{2}{c}{$k=50$} & \multicolumn{2}{c}{$k=20$} & \multicolumn{2}{c}{$k=100$} & \multicolumn{2}{c}{$k=20$} & \multicolumn{2}{c}{$k=50$}  & \multicolumn{2}{c}{$k=20$} & \multicolumn{2}{c}{$k=50$} \\
\cmidrule(r){2-3} \cmidrule(r){4-5} \cmidrule(lr){6-7} \cmidrule(lr){8-9} \cmidrule(lr){10-11} \cmidrule(l){12-13} \cmidrule(lr){14-15} \cmidrule(l){16-17}
\textbf{Method} & Coverage & Time (s) & Coverage & Time (s) & Coverage & Time (s) & Coverage & Time (s) & Coverage & Time (s) & Coverage & Time (s) & Coverage & Time (s) & Coverage & Time (s) \\
\midrule
Random (240s) & 13488.87 & 240.00019 & 36912.84 & 240.00020 & 24438.53 & 240.00009 & 70717.71 & 240.00012 & 5287.67 & 240.00103 & 7367.00 & 240.00089 & 13898.00 & 240.00028 & 16036.50 & 240.00022 \\
Gurobi (120s) & 15714.90 & 0.93991 & 44972.28 & 120.07712 & 31347.62 & 21.62572 & 89702.37 & 120.05282 & 5640.00 & 120.72583 & 7610.00 & 120.62132 & 25864.00 & 0.56769 & 30560.83 & 0.53344 \\
Greedy & 15633.93 & 0.00721 & 44312.81 & 0.02072 & 31132.96 & 0.02318 & 88698.89 & 0.07883 & 5617.00 & 0.07212 & 7630.00 & 0.13551 & 25855.33 & 0.06675 & 30542.17 & 0.12559 \\
Ucom2-short & 15471.03 & 0.45178 & 44614.61 & 0.42560 & 30016.16 & 0.84422 & 89102.86 & 0.84090 & 5525.00 & 1.42266 & 7634.00 & 1.21777 & 25833.00 & 8.24705 & 30526.00 & 7.75699 \\
\midrule
GeoNCO & 15146.49 & 0.96187 & 40984.63 & 2.23917 & 29878.15 & 1.28572 & 81428.14 & 3.72534 & 5343.00 & 2.68297 & 7411.67 & 4.04408 & 8350.33 & 2.44298 & 11351.67 & 3.21277 \\
CardNN-noTTO-S & 9231.54 & 1.33321 & 33051.73 & 1.32723 & 18458.92 & 1.38980 & 65788.15 & 1.35915 & 5074.33 & 0.94137 & 7193.00 & 0.99340 & 77.00 & 0.66294 & 173.00 & 0.62554 \\
EGN-naive & 15262.76 & 53.041 & 41272.68 & 120.136 & 29968.04 & 120.010 & 81166.12 & 120.356 & 5234.67 & 120.676 & 7408.33 & 121.335 & 152.67 & 108.30300 & 247.17 & 105.46050 \\
FWNCO & 15161.20 & 0.36310 & 42087.45 & 0.35790 & 29807.93 & 0.39220 & 83343.16 & 0.57820 & 5344.67 & 1.00100 & 7393.00 & 1.27860 & 8332.00 & 0.68350 & 11042.50 & 0.81810 \\
\midrule
Ours & 15056.48 & 0.00383 & 42165.90 & 0.00353 & 29932.85 & 0.00350 & 84039.51 & 0.00352 & 5318.33 & 0.00846 & 7393.20 & 0.01191 & 10095.40 & 0.00910 & 16276.30 & 0.00842 \\
\bottomrule
\end{tabular}%
}
\end{table}

\section{Details and Further Experiments on Densest $k$-Subgraph}
\label{sec:appendix-dks}

\subsection{Experiment Setup}
\label{app:dks-setup}

\paragraph{Training and test graphs.}
We train on synthetic graphs with 500 vertices and evaluate transfer to
real-world networks. The training graphs are randomly generated with varying
edge densities and densely connected groups of vertices. We train a single model with
$k\in\{20,50,100,200\}$, sampling one value of $k$ for each training batch.
Testing uses the 12 graphs from the SNAP dataset collection~\citep{leskovec2016snap} in
Table~\ref{tab:dks-completed}, combining the graph collections used by
FW-DL~\citep{lu2025densest} and EP-Prox~\citep{liu2026scalable}.
They include social, collaboration, web, Internet, and road networks,
up to approximately 4.85 million vertices. All tests use $k=100$ and count each induced undirected edge once. The main table normalizes each count by the highest reported count on the same graph. Raw counts are provided in
Table~\ref{tab:dks-raw-edges}.

\paragraph{Graph representation.}
To accommodate DKS instances with millions of vertices, we use compact
node-level features to encode graph information and a shared MLP to generate
the query vectors. The features and query vectors are computed in chunks
on the sparse graph, limiting peak GPU memory usage. This illustrates the
scalability of our framework: the graph representation and neural network
can be adapted to the problem size while retaining the same optimization
procedure.

To construct the inputs used to generate $m$ query vectors, we sample
vertices $a_1,\ldots,a_m$ without replacement from pools ranked by core number and degree. For each sampled vertex $a_j$, its closed
neighborhood $N[a_j]=\{a_j\}\cup N(a_j)$ describes the surrounding graph structure. For each vertex $u$ in the entire graph, we compute 10 features describing its degree and core number, membership in $N[a_j]$, normalized neighborhood-overlap counts, and its core number relative to that of $a_j$. This gives a feature matrix $H_j\in\mathbb{R}^{n\times10}$ for each sampled
vertex $a_j$, where row $u$ describes graph vertex $u$ relative to $a_j$.

\paragraph{Neural Network Training and inference.}
We use one shared MLP with 10 input features, hidden-layer widths 32 and 16, and SiLU activations to produce m query vectors $q_j$.
At training, we train for 30 epochs with 128 new graphs per epoch, batch size 4, and $m_{\mathrm{train}}=32$. AdamW uses learning rate $10^{-3}$, weight decay $10^{-5}$, and gradient clipping at norm 1.
Checkpoint selection uses mean density across the four budgets on 24
synthetic validation graphs with 1000 vertices.
At inference, we generate $m$ query vectors, apply top-$k$, score
the resulting subsets, and return the best one. We use query chunks of 16 on the full sparse graph, processing 16 independent queries at a time. Runtimes include graph/model preparation and inference.

\subsection{Related Work and Baselines}
\label{app:dks-baselines}

We compare with three continuous-optimization approaches.
FW-DL~\citep{lu2025densest} applies Frank--Wolfe to a diagonally loaded
quadratic relaxation. L-ADMM~\citep{konar2021exploring} optimizes a Lov\'asz-extension
relaxation using ADMM. EP-Prox~\citep{liu2026scalable} uses an exact-penalty
reformulation with a proximal-gradient algorithm.
These methods optimize each test graph directly. Our neural network is
trained on synthetic graphs and its parameters remain fixed during testing.

We use FW-DL and L-ADMM from the public DkS-Diagonal-Loading
implementation provided by the authors\footnote{\url{https://github.com/luqh357/DkS-Diagonal-Loading}},
with the released demo settings of 200 and 2000 iterations, respectively.
EP-Prox uses the authors' MATLAB implementation and public demo defaults.
All methods are evaluated on the same prepared graph structures at $k=100$,
and objectives are reported as induced-edge counts.
For Orkut, input conversion for FW-DL/L-ADMM is performed in blocks because
of memory limits; this conversion time is excluded for fair comparison,
with the optimizer unchanged.

\subsection{Effect of the Parallel Query Budget}
\label{app:dks-query-budget}
During training, we use a fixed query budget $m_{\mathrm{train}}=32$.
At inference, we keep the same $m_{\mathrm{train}}=32$
checkpoint and evaluate $m\in\{100,200,500,1000\}$.
The neural network shares its parameters across all $m$ queries.

Table~\ref{tab:dks-m-ablation} shows that the mean induced-edge count increases
from 3016.08 at $m=100$ to 3135.42 at $m=1000$, while mean runtime rises from
3.5297 to 13.2186 seconds. Larger budgets are particularly useful on
com-Youtube and soc-Pokec, where the counts increase from 2035 to 2397 and
from 2064 to 2617, respectively. Extremely large graphs such as com-DBLP and
soc-LiveJournal1 already attain the maximum possible 4950 edges at $m=100$.

\begin{table}[H]
\centering
\caption{Effect of the parallel query budget $m$ on DKS ($k=100$), using one model trained with $m=32$.}
\label{tab:dks-m-ablation}
\resizebox{\linewidth}{!}{\begin{tabular}{lrrrrrrrr}
\toprule
Graph & \multicolumn{2}{c}{$m=100$} & \multicolumn{2}{c}{$m=200$} & \multicolumn{2}{c}{$m=500$} & \multicolumn{2}{c}{$m=1000$} \\
 & Edges & Time (s) & Edges & Time (s) & Edges & Time (s) & Edges & Time (s) \\
\midrule
Facebook & 4789.00 & 0.8004 & 4802.00 & 0.5067 & 4805.00 & 0.7214 & 4819.00 & 0.5134 \\
web-Stanford & 4149.00 & 0.8109 & 4149.00 & 0.7799 & 4143.00 & 1.1586 & 4143.00 & 1.5178 \\
com-Youtube & 2035.00 & 1.2997 & 2117.00 & 1.5544 & 2216.00 & 2.7178 & 2397.00 & 4.6347 \\
soc-Pokec & 2064.00 & 3.1761 & 2212.00 & 3.6994 & 2508.00 & 5.4008 & 2617.00 & 9.1007 \\
as-Skitter & 4256.00 & 1.9902 & 4256.00 & 2.5392 & 4471.00 & 4.2418 & 4479.00 & 7.1603 \\
com-Orkut & 3274.00 & 15.0157 & 3404.00 & 16.0264 & 3413.00 & 19.6053 & 3424.00 & 27.5313 \\
ca-HepTh & 1160.00 & 0.5066 & 1160.00 & 0.4960 & 1160.00 & 0.4866 & 1160.00 & 0.5276 \\
ca-CondMat & 1022.00 & 0.4532 & 1049.00 & 0.4883 & 1054.00 & 0.5487 & 1027.00 & 0.5730 \\
com-DBLP & 4950.00 & 0.6359 & 4950.00 & 0.7666 & 4950.00 & 1.0709 & 4950.00 & 1.5621 \\
roadNet-CA & 40.00 & 1.5208 & 43.00 & 2.1781 & 43.00 & 4.2050 & 46.00 & 7.4008 \\
wiki-Talk & 3504.00 & 1.7402 & 3553.00 & 2.5712 & 3541.00 & 5.0471 & 3613.00 & 9.0202 \\
soc-LiveJournal1 & 4950.00 & 14.4073 & 4950.00 & 22.4427 & 4950.00 & 48.1916 & 4950.00 & 89.0817 \\
Mean & 3016.08 & 3.5297 & 3053.75 & 4.5041 & 3104.50 & 7.7829 & 3135.42 & 13.2186 \\
\bottomrule
\end{tabular}}
\end{table}

\subsection{Raw Results}
Table~\ref{tab:dks-raw-edges} provides the induced-edge counts used in Table~\ref{tab:dks-completed}. Normalization divides each count by the largest count on the same graph; the main-table Mean averages the 12 resulting ratios.

\begin{table*}[t]
\centering
\caption{DKS results on 12 SNAP graphs ($k=100$). Normalized objective: higher is better; time: lower is better. \textcolor{Red!50}{$\blacksquare$} and \textcolor{Blue!50}{$\blacksquare$} indicate the best and second-best results, respectively.}
\label{tab:dks-completed}
\setlength{\tabcolsep}{3pt}
\resizebox{\textwidth}{!}{%
\begin{tabular}{lrrrrrrrrrrrr}
\toprule
& \multicolumn{2}{c}{Gurobi (120s)} & \multicolumn{2}{c}{FW-DL} & \multicolumn{2}{c}{L-ADMM} & \multicolumn{2}{c}{EP-Prox} & \multicolumn{2}{c}{Ours ($m=100$)} & \multicolumn{2}{c}{Ours ($m=1000$)} \\
\cmidrule(lr){2-3}\cmidrule(lr){4-5}\cmidrule(lr){6-7}\cmidrule(lr){8-9}\cmidrule(lr){10-11}\cmidrule(lr){12-13}
\textbf{Dataset} & Norm. obj. $\uparrow$ & Time (s) $\downarrow$ & Norm. obj. $\uparrow$ & Time (s) $\downarrow$ & Norm. obj. $\uparrow$ & Time (s) $\downarrow$ & Norm. obj. $\uparrow$ & Time (s) $\downarrow$ & Norm. obj. $\uparrow$ & Time (s) $\downarrow$ & Norm. obj. $\uparrow$ & Time (s) $\downarrow$ \\
\midrule
\textbf{Facebook} & \rkone{1.00} & 118.0970 & \rktwo{1.00} & 0.5495 & 0.99 & 4.0285 & 0.99 & \rkone{0.1015} & 0.98 & 0.8004 & 0.99 & \rktwo{0.5134} \\
\textbf{web-Stanford} & 0.00 & 120.0404 & \rkone{1.00} & 11.1951 & 0.29 & 148.3811 & 0.06 & \rktwo{0.8216} & \rktwo{1.00} & \rkone{0.8109} & 1.00 & 1.5178 \\
\textbf{com-Youtube} & 0.06 & 120.0634 & \rkone{1.00} & 21.9967 & \rktwo{0.94} & 411.8932 & 0.50 & \rkone{1.0469} & 0.77 & \rktwo{1.2997} & 0.90 & 4.6347 \\
\textbf{soc-Pokec} & 0.12 & 120.0375 & 0.52 & 162.3344 & 0.30 & 1880.5188 & 0.17 & \rktwo{5.7141} & \rktwo{0.79} & \rkone{3.1761} & \rkone{1.00} & 9.1007 \\
\textbf{as-Skitter} & 0.02 & 120.0398 & \rkone{1.00} & 52.7224 & 0.94 & 1143.8981 & 0.53 & \rktwo{4.8766} & 0.93 & \rkone{1.9902} & \rktwo{0.97} & 7.1603 \\
\textbf{com-Orkut} & 0.10 & 121.1402 & \rkone{1.00} & 62.4983 & 0.74 & 8644.3508 & 0.75 & \rktwo{17.3336} & 0.84 & \rkone{15.0157} & \rktwo{0.88} & 27.5313 \\
\textbf{ca-HepTh} & \rkone{1.00} & 118.0885 & 0.78 & \rktwo{0.3836} & 1.00 & 3.0927 & 0.76 & \rkone{0.1042} & \rktwo{1.00} & 0.5066 & \rktwo{1.00} & 0.5276 \\
\textbf{ca-CondMat} & \rkone{1.00} & 25.9802 & 0.90 & 0.9467 & \rktwo{0.99} & 7.7723 & 0.80 & \rkone{0.1322} & 0.81 & \rktwo{0.4532} & 0.81 & 0.5730 \\
\textbf{com-DBLP} & \rktwo{0.01} & 120.0482 & \rkone{1.00} & 6.7888 & \rkone{1.00} & 102.7746 & \rkone{1.00} & \rkone{0.4909} & \rkone{1.00} & \rktwo{0.6359} & \rkone{1.00} & 1.5621 \\
\textbf{roadNet-CA} & \rktwo{0.59} & 118.9115 & 0.19 & 27.9232 & \rkone{1.00} & 538.4673 & 0.13 & \rktwo{2.4999} & 0.22 & \rkone{1.5208} & 0.25 & 7.4008 \\
\textbf{wiki-Talk} & 0.11 & 118.9754 & \rkone{1.00} & 33.8185 & \rktwo{0.97} & 1014.6148 & 0.97 & \rktwo{1.8537} & 0.90 & \rkone{1.7402} & 0.93 & 9.0202 \\
\textbf{soc-LiveJournal1} & 0.04 & 120.0782 & 0.63 & 326.1160 & \rktwo{1.00} & 4861.7574 & 1.00 & \rkone{12.5898} & \rkone{1.00} & \rktwo{14.4073} & \rkone{1.00} & 89.0817 \\
\midrule
\textbf{Mean} & 0.34 & 111.7917 & 0.83 & 58.9394 & 0.85 & 1563.4625 & 0.64 & \rktwo{3.9638} & \rktwo{0.85} & \rkone{3.5297} & \rkone{0.89} & 13.2186 \\
\bottomrule
\end{tabular}}
\par\smallskip
\begin{minipage}{\textwidth}\footnotesize
Normalized objective is the induced-edge count divided by the highest count in the same row. Mean averages these ratios over all 12 graphs. Raw counts are provided in Table~\ref{tab:dks-raw-edges}.
Orkut input conversion for FW-DL/L-ADMM was performed in blocks due to out-of-memory errors; conversion time is excluded for fair comparison.
\end{minipage}
\end{table*}

\begin{table}[H]
\centering
\caption{Raw DKS induced-edge counts and inference time on 12 SNAP graphs ($k=100$). Edges: higher is better; time: lower is better.}
\label{tab:dks-raw-edges}
\setlength{\tabcolsep}{3pt}
\resizebox{\linewidth}{!}{%
\begin{tabular}{lrrrrrrrrrrrr}
\toprule
& \multicolumn{2}{c}{Gurobi (120s)} & \multicolumn{2}{c}{FW-DL} & \multicolumn{2}{c}{L-ADMM} & \multicolumn{2}{c}{EP-Prox} & \multicolumn{2}{c}{Ours ($m=100$)} & \multicolumn{2}{c}{Ours ($m=1000$)} \\
\cmidrule(lr){2-3}\cmidrule(lr){4-5}\cmidrule(lr){6-7}\cmidrule(lr){8-9}\cmidrule(lr){10-11}\cmidrule(lr){12-13}
Dataset & Edges & Time (s) & Edges & Time (s) & Edges & Time (s) & Edges & Time (s) & Edges & Time (s) & Edges & Time (s) \\
\midrule
Facebook & 4872 & 118.0970 & 4871 & 0.5495 & 4824 & 4.0285 & 4837 & 0.1015 & 4789 & 0.8004 & 4819 & 0.5134 \\
web-Stanford & 0 & 120.0404 & 4151 & 11.1951 & 1190 & 148.3811 & 256 & 0.8216 & 4149 & 0.8109 & 4143 & 1.5178 \\
com-Youtube & 156 & 120.0634 & 2659 & 21.9967 & 2496 & 411.8932 & 1334 & 1.0469 & 2035 & 1.2997 & 2397 & 4.6347 \\
soc-Pokec & 308 & 120.0375 & 1363 & 162.3344 & 796 & 1880.5188 & 436 & 5.7141 & 2064 & 3.1761 & 2617 & 9.1007 \\
as-Skitter & 106 & 120.0398 & 4595 & 52.7224 & 4341 & 1143.8981 & 2449 & 4.8766 & 4256 & 1.9902 & 4479 & 7.1603 \\
com-Orkut & 373 & 121.1402 & 3889 & 62.4983 & 2884 & 8644.3508 & 2929 & 17.3336 & 3274 & 15.0157 & 3424 & 27.5313 \\
ca-HepTh & 1164 & 118.0885 & 908 & 0.3836 & 1159 & 3.0927 & 888 & 0.1042 & 1160 & 0.5066 & 1160 & 0.5276 \\
ca-CondMat & 1267 & 25.9802 & 1135 & 0.9467 & 1251 & 7.7723 & 1014 & 0.1322 & 1022 & 0.4532 & 1027 & 0.5730 \\
com-DBLP & 57 & 120.0482 & 4950 & 6.7888 & 4950 & 102.7746 & 4950 & 0.4909 & 4950 & 0.6359 & 4950 & 1.5621 \\
roadNet-CA & 106 & 118.9115 & 34 & 27.9232 & 181 & 538.4673 & 23 & 2.4999 & 40 & 1.5208 & 46 & 7.4008 \\
wiki-Talk & 426 & 118.9754 & 3899 & 33.8185 & 3800 & 1014.6148 & 3768 & 1.8537 & 3504 & 1.7402 & 3613 & 9.0202 \\
soc-LiveJournal1 & 187 & 120.0782 & 3126 & 326.1160 & 4948 & 4861.7574 & 4944 & 12.5898 & 4950 & 14.4073 & 4950 & 89.0817 \\
\midrule
Mean & 751.83 & 111.7917 & 2965.00 & 58.9394 & 2735.00 & 1563.4625 & 2319.00 & 3.9638 & 3016.08 & 3.5297 & 3135.42 & 13.2186 \\
\bottomrule
\end{tabular}}
\par\smallskip
\begin{minipage}{\linewidth}\footnotesize
Orkut input conversion for FW-DL/L-ADMM was performed in blocks due to out-of-memory errors; conversion time is excluded for fair comparison. The original baseline implementations are unchanged.
\end{minipage}
\end{table}

\subsection{Ablation on Training}
\label{app:dks-training-ablation}

Since the input features are a strong signal of the underlying structure, we investigate the impact of training on the model's performance. We compare the trained checkpoint with its saved initialization, retaining
the same features, fixed terms in $q_j$, sampled vertices, top-$k$
oracle, and best-candidate selection. The initial neural residual is zero,
so the untrained model uses the two fixed terms in $q_j$.
Trained and untrained models use identical sampled vertices within each
repetition. We report results at
$m=200$ and $m=1000$.

Table~\ref{tab:dks-learning-ablation} shows that training increases the mean
edge count from 2324.83 to 3043.86 at $m=200$, and from 2484.22 to 3127.00
at $m=1000$, corresponding to improvements of 30.9\% and 25.9\%.
Large gains on com-Youtube, as-Skitter, com-Orkut, and wiki-Talk demonstrate that learning improves upon vertex sampling and the fixed features.

\begin{table}[H]
\small
\centering
\caption{Effect of training on Ours for DKS ($k=100$, training $m=32$). Mean induced edges over three repetitions; higher is better.}
\label{tab:dks-learning-ablation}
% \setlength{\tabcolsep}{4pt}
% \resizebox{\linewidth}{!}{%
\begin{tabular}{lrrrr}
\toprule
& \multicolumn{2}{c}{Test $m=200$} & \multicolumn{2}{c}{Test $m=1000$} \\
\cmidrule(lr){2-3}\cmidrule(lr){4-5}
Dataset & Untrained & Trained & Untrained & Trained \\
\midrule
Facebook & 4229.00 & 4804.67 & 4425.33 & 4820.67 \\
web-Stanford & 4134.67 & 4143.67 & 4143.00 & 4144.33 \\
com-Youtube & 841.33 & 2184.33 & 1067.67 & 2341.67 \\
soc-Pokec & 1409.33 & 2132.00 & 1806.67 & 2577.33 \\
as-Skitter & 2692.33 & 4275.67 & 2866.67 & 4477.00 \\
com-Orkut & 2080.33 & 3356.67 & 2296.67 & 3425.33 \\
ca-HepTh & 577.33 & 1160.33 & 622.67 & 1161.33 \\
ca-CondMat & 780.33 & 1032.00 & 912.33 & 1070.00 \\
com-DBLP & 4950.00 & 4950.00 & 4950.00 & 4950.00 \\
roadNet-CA & 114.00 & 45.00 & 115.33 & 49.67 \\
wiki-Talk & 1139.33 & 3492.00 & 1654.33 & 3556.67 \\
soc-LiveJournal1 & 4950.00 & 4950.00 & 4950.00 & 4950.00 \\
\midrule
Mean (12 graphs) & 2324.83 & 3043.86 & 2484.22 & 3127.00 \\
\bottomrule
\end{tabular}
% }
\end{table}

% \color{red}
\section{Ablation on aggregation}
\label{sec:appendix-aggregation}

Section~\ref{sec:method} describes how query generation and aggregation
produce the continuous representation $\bx_\theta$ used for training. Simple
averaging achieves strong empirical results across our benchmarks
without additional complexity, but the framework also permits direct
query generation and alternative aggregation rules. Here, we examine
the role of Gaussian sampling and evaluate learnable aggregation as
an alternative that introduces query-dependent gradient paths.

To further assess these design choices, we conducted additional
controlled experiments examining the effects of removing Gaussian
sampling and changing the aggregation rule. Specifically, we compared
Gaussian versus direct query generation and mean versus learnable
aggregation. We include the Gaussian-free variant to test whether the
learned query generator itself can maintain candidate diversity without
relying on injected Gaussian noise.

Within each problem, we retained the original training seed, training
data, training budget, and candidate budget. MC uses Random1000 with
$k=20$ and $m=50$; QAP uses the 108 QAPLIB instances with $m=5000$.

For learnable aggregation, a shared scoring network assigns a scalar
score to each query based on its normalized statistics. Softmax converts
these scores into weights, giving for $Q=(\bq_1,\ldots,\bq_m)$
\[
    \bx_\theta = \sum_{i=1}^{m} w_i(Q)\bq_i.
\]
The scorer adds 96 trainable parameters and is initialized to uniform
weights, so training starts from mean aggregation. It is optimized
jointly with the query generator, introducing query-dependent gradient
paths through both the weights and the queries.

\begin{table}[t]
\small
    \centering
    \caption{\rev{Comparison of query generation and aggregation rules
    on MC and QAP.}}
    \label{tab:query-aggregation-ablation}
    \begin{tabular}{llrr}
        \toprule
        Query generation & Aggregation
        & \shortstack{MC coverage\\$\uparrow$}
        & \shortstack{QAP mean class\\gap (\%) $\downarrow$} \\
        \midrule
        Gaussian & Mean    & 29932.85 & 26.39 \\
        Gaussian & Learned & \textbf{30045.10} & \textbf{25.84} \\
        Direct   & Mean    & 29950.36 & 26.61 \\
        \bottomrule
    \end{tabular}
\end{table}

Learnable aggregation improves MC coverage by $0.375\%$ and reduces
the QAP gap by $0.550$ percentage points relative to mean aggregation.
Direct query generation achieves slightly higher MC coverage than
Gaussian generation, while Gaussian sampling achieves a lower QAP gap
by $0.220$ percentage points. These results demonstrate the practicality
of alternative aggregations and show that the contribution of Gaussian
exploration depends on the problem.

We favor Gaussian query generation for its scalability and flexible
candidate budget. Once the network predicts the distribution parameters,
additional queries can be generated through parallel sampling while
reusing those parameters and the encoded instance representation. This
allows the candidate budget to grow without adding a separate learned
output for every additional query, making the approach well suited to
large instances and large candidate pools. Fresh samples also provide
additional opportunities to explore different polytope regions.

We retain mean aggregation as a practical trade-off between solution
quality and training cost. Mean aggregation requires only a simple
reduction, whereas learnable aggregation adds query scoring, softmax
weighting, and backpropagation through the scoring network. Given the
modest quality gains observed here, mean aggregation provides an
efficient default, while learnable aggregation offers an option when
additional training computation is acceptable.
% \color{red}
\section{Residual and Quality of Generated Solutions}
\label{sec:residual}

Our approach constructs a probability distribution over feasible
candidates generated in parallel, without requiring an exact
decomposition of $\bx_\theta$. The coefficients remain valid
probabilities since $\balpha(\bx_\theta)\in\Delta_m$, regardless of
whether $\bx_\theta$ lies inside $\operatorname{conv}(V)$.
Moreover, \Cref{prop:aediff} does not require the projection residual
$\|\bx_\theta-V\balpha(\bx_\theta)\|_2$ to be small.

\paragraph{Interpretation of the projection residual.}
The projection residual is not necessarily indicative of solution
quality. For an exact LMO with fixed tie-breaking among maximizers,
scaling all queries by $c>0$ leaves the generated candidates unchanged:
\[
    \operatorname{LMO}(c\bq_i)
    = \operatorname{LMO}(\bq_i).
\]
Under mean aggregation, this scaling replaces $\bx_\theta$ by
$c\bx_\theta$. For $\bx_\theta\neq\mathbf{0}$, the residual
$\|c\bx_\theta-V\balpha(c\bx_\theta)\|_2$ can grow arbitrarily large. Conversely, a small residual
does not guarantee high-quality candidates. Our method directly
optimizes a surrogate over the generated solutions during training
and selects the best feasible candidate at inference.

An aggregate $\bx_\theta$ far from the candidate hull can still
correspond to queries that lead the LMO to high-quality feasible
solutions. Likewise, an aggregate close to or inside the hull can
correspond to solutions with different objective values. The geometric
position of $\bx_\theta$ therefore provides limited information about
the optimization outcomes associated with the queries.

\paragraph{Quality of the generated solutions during training.}
To assess whether training guides the queries toward better feasible
solutions, we measure the mean objective value of the LMO solutions
corresponding to queries $\bq_1,\ldots,\bq_m$:
\[
    \bar f_{\mathrm{LMO}}
    = \frac{1}{m}\sum_{i=1}^{m}
      f\bigl(\operatorname{LMO}(\bq_i)\bigr).
\]
Each query contributes equally to this average. This statistic directly
measures the average objective quality of the feasible solutions
generated by the query set. We also plot our combinatorial loss,
defined in Eq.~\eqref{eq:extension}, alongside this unweighted mean.
Improvements in the mean LMO objective therefore provide empirical
evidence that the training updates are useful, even when the projection
residual remains large or fluctuates.

\begin{figure}[t]
    \centering
    \includegraphics[width=\linewidth]{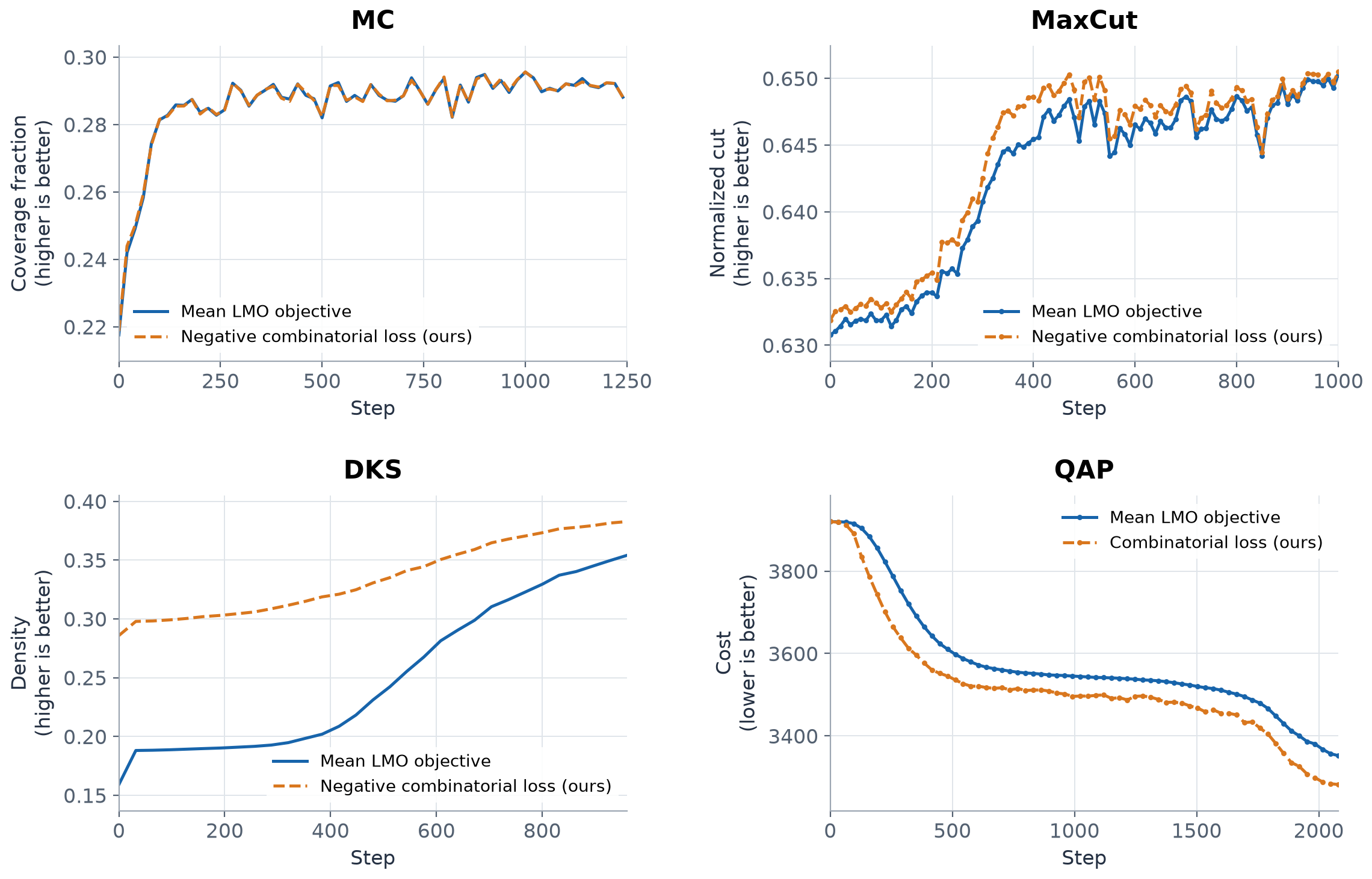}
    \caption{\rev{Mean objective value of the LMO solutions and our
    combinatorial loss throughout training. For MC, MaxCut, and DkS,
    the negative loss is plotted, so higher values indicate better
    solutions. For QAP, lower values indicate better solutions.}}
    \label{fig:training-lmo-objectives}
\end{figure}

\begin{figure}[t]
    \centering
    \includegraphics[width=\linewidth]{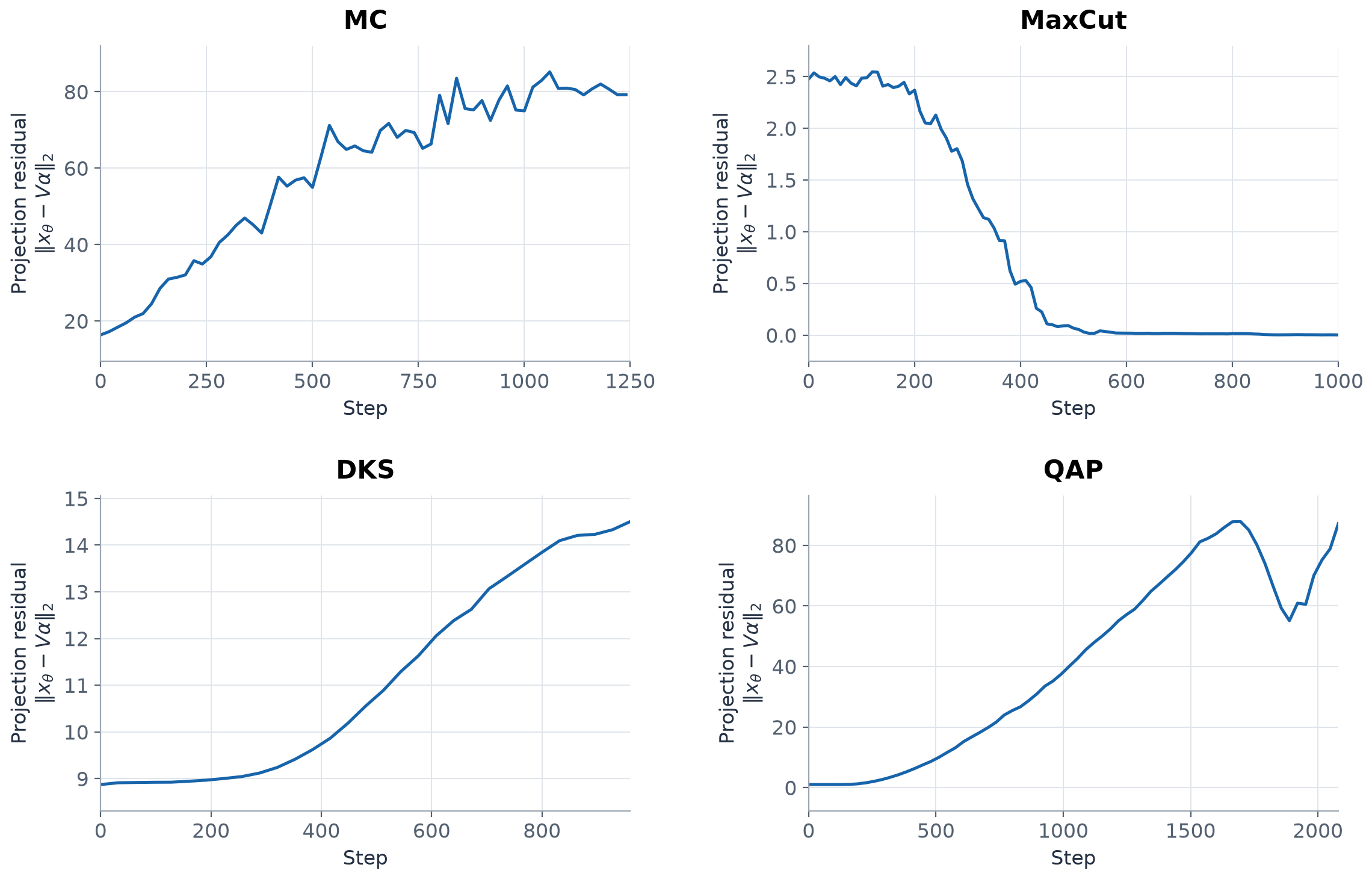}
    \caption{\rev{Projection residual
    $\|\bx_\theta-V\balpha(\bx_\theta)\|_2$
    throughout training on MC, MaxCut, DkS, and QAP.}}
    \label{fig:training-projection-residuals}
\end{figure}

Figures~\ref{fig:training-lmo-objectives}
and~\ref{fig:training-projection-residuals} track the mean objective
value of the LMO solutions returned for all queries, our combinatorial
loss, and the projection residual throughout training. For MC, MaxCut,
and DkS, we plot the negative loss so that both quality curves follow
the ``higher is better'' convention; for QAP, we plot the loss itself,
with lower values indicating better solutions.

\begin{itemize}
    \item \textbf{MaxCut:}
    The mean normalized cut and the negative combinatorial loss
    improve overall, while the projection residual decreases
    toward zero.
    
    \item \textbf{MC:}
    Mean coverage and the negative combinatorial loss increase and
    then stabilize, while the projection residual increases before
    leveling off.

    \item \textbf{DkS:}
    Mean density and the negative combinatorial loss increase,
    while the projection residual also increases overall.

    \item \textbf{QAP:}
    Both mean cost and the combinatorial loss decrease, while
    the projection residual shows a substantial overall increase.
\end{itemize}

The mean LMO objective measures the quality of the feasible solutions
generated by all queries, giving each query equal weight. Its improvement
therefore provides evidence that training guides the queries toward
better solutions beyond improving their $\balpha$-weighted objective
alone. Across these experiments, this improvement occurs under different
residual behaviors, supporting the effectiveness of the training signal.

% These observations are consistent with Proposition~1, which applies to aggregates both inside and outside the polytope without requiring a small projection residual.
\end{document}